\documentclass{article}

\usepackage[final]{neurips_2024}

\usepackage[utf8]{inputenc} 
\usepackage[T1]{fontenc}    
\usepackage{hyperref}       
\usepackage{url}            
\usepackage{booktabs}       
\usepackage{amsfonts}       
\usepackage{nicefrac}       
\usepackage{microtype}      
\usepackage{xcolor}         

\usepackage{algorithm}
\usepackage{algpseudocode}
\usepackage{amsmath}
\DeclareMathOperator*{\argmax}{arg\,max}
\usepackage{multirow}

\usepackage{geometry}
\usepackage{graphicx} 
\usepackage{subcaption}

\title{Generative Model for Synthesizing Ionizable Lipids: A Monte Carlo Tree Search Approach}

\author{%
  Jingyi Zhao\\
  Department of Engineering\\
  University of Cambridge\\
  Cambridge, CB2 1PZ\\
  \texttt{jz610@cam.ac.uk} \\
  \And
  Yuxuan Ou\\
  Department of Engineering\\
  University of Cambridge\\
  Cambridge, CB2 1PZ\\
  \texttt{yuxuan.ou@trinity.ox.ac.uk} \\
  \And
  Austin Tripp\\
  Department of Engineering\\
  University of Cambridge\\
  Cambridge, CB2 1PZ\\
  \texttt{ajt212@cam.ac.uk} \\
  \And
  Morteza Rasoulianboroujeni\\
  School of Pharmacy\\
  University of Wisconsin-Madison\\
  Madison, WI 53705\\
  \texttt{rasoulianbor@wisc.edu} \\
  \And
  José Miguel Hernández-Lobato\\
  Department of Engineering\\
  University of Cambridge\\
  Cambridge, CB2 1PZ\\
  \texttt{jmh233@cam.ac.uk} \\
}

\begin{document}

\maketitle

\begin{abstract}
Ionizable lipids are essential in developing lipid nanoparticles (LNPs) for effective messenger RNA (mRNA) delivery. While traditional methods for designing new ionizable lipids are typically time-consuming, deep generative models have emerged as a powerful solution, significantly accelerating the molecular discovery process. However, a practical challenge arises as the molecular structures generated can often be difficult or infeasible to synthesize. This project explores Monte Carlo tree search (MCTS)-based generative models for synthesizable ionizable lipids. Leveraging a synthetically accessible lipid building block dataset and two specialized predictors to guide the search through chemical space, we introduce a policy network guided MCTS generative model capable of producing new ionizable lipids with available synthesis pathways.
\end{abstract}

\section{Introduction}
The development of messenger RNA (mRNA)-based therapeutics marks a transformative advance in the treatment and prevention of a wide range of diseases, including genetic disorders, infectious diseases, and cancer \citep{mrna-based_2014, mRNA-KONG2023114993}. Given the intrinsic instability of mRNA, a robust delivery mechanism is crucial, with ionizable lipid nanoparticles (LNPs) emerging as the leading technology for this purpose \citep{nanoparticles-for-drug-delivery, LNP}. These LNPs comprise four distinct lipid types, among which the ionizable amine-containing lipids are pivotal. They primarily facilitate the encapsulation of mRNA within LNPs and enhance its delivery into the cellular cytoplasm, where it can be translated into therapeutic proteins \citep{ionizable-lipid-1, ionizable-lipid-2, ionizable-lipid-3, ionizable-lipid-4}. Notably, different LNPs often feature unique ionizable lipid structures, emphasizing the importance of developing a diverse array of ionizable lipids. This diversity is crucial for effectively delivering mRNA to various target cells and tissues, highlighting the need for continued innovation in the design of ionizable lipids to facilitate mRNA-based therapeutics.

Designing novel ionizable lipids is traditionally time-consuming and labor-intensive. However, recent advancements in the integration of deep learning with combinatorial chemistry have shown great promise in accelerating this development \citep{AGILE}. One of the primary challenges in ionizable lipid generation is the effective exploration of a vast combinatorial chemical space. The SyntheMol approach has demonstrated considerable success in this area, utilizing Monte Carlo tree search (MCTS) to efficiently navigate through expansive search spaces and generate desired molecules, complete with synthesis pathways, particularly in the field of antibiotic development \citep{SyntheMol}.

This work extends the application of the MCTS-based generative model to the domain of ionizable lipid generation. We introduce a policy network guided MCTS method which leverages the strengths of MCTS and integrates it with the strategic direction provided by the policy network, aiming to optimize the generation process and yield new ionizable lipids more efficiently. Our main contributions are:
\begin{itemize}
    \item We compile a synthetically accessible dataset of molecules suitable for use as ionizable lipid heads or lipid tails, which can facilitate future ionizable lipid generation tasks.
    \item We develop reliable lipid property predictors to assess whether a candidate molecule possesses lipid-like characteristics or ionizable properties.
    \item We propose and implement a generative model that leverages a policy network to guide the MCTS in exploring the chemical space. This model effectively generates high-quality products, complete with available synthesis pathways.
\end{itemize}

\section{Related Work}

Deep generative models have emerged as a powerful solution to the inverse molecular design challenge, enabling the translation of desired molecular properties into specific molecular structures \citep{GVAE, CVAE, GraphVAE, GAN-DEFactor, dcGAN, Diffusion-1, Diffusion-2}. A practical problem that obstructs the usefulness of these generative algorithms is that proposed molecular structures may be challenging or infeasible to synthesize \citep{gao2020synthesizability}. While post-hoc synthesis planning for generated molecules is feasible \citep{gao2020synthesizability, AlphaChem, segler_planning_2018}, a more effective approach is to incorporate synthesis instructions directly into the design phase. One effective solution is to adopt a bottom-up approach which begins with existing building blocks and strategically determines pathways to synthesize product molecules possessing desired properties \citep{DOG, SyntheMol}.

Identifying new ionizable lipids remains a bottleneck of LNP development. Current state-of-the-art approaches still primarily rely on combinatorial chemistry techniques. Even when synthesized, an ionizable lipid often fails to exhibit transfection capabilities, with a low likelihood of achieving high transfection efficiency \citep{AGILE, Bowen-TE}.  Despite the vast structural design space of ionizable lipids, small modifications in chemical structure can lead to substantial differences in biological performance \citep{Bowen-TE}. Existing approaches that incorporate machine learning into ionizable lipid development primarily focus on building transfection efficiency predictors to aid in lipid screening \citep{AGILE, Bowen-TE, ding2023-TE, TE-LLM}. However, no current approaches directly utilize machine learning for the generative design of ionizable lipids.

\section{Background}
This section presents the theoretical background of this work.

\subsection{Monte Carlo Tree Search in Molecular Generation}
MCTS is a robust algorithm that integrates the stochastic nature of Monte Carlo simulations with the structured decision-making processes inherent in tree searches \citep{UCT, Coulom2006EfficientSA}. A recent study has shown that MCTS-based generative models can perform successfully in small-molecule antibiotic development \citep{SyntheMol}. In the context of SyntheMol, the algorithm iteratively builds a tree structure, where each node represents one or more potential molecular structures, and branches represent possible synthesis steps using 13 well-validated chemical reactions \citep{SyntheMol}. The application of SyntheMol to lipid generation is detailed in Appendix \ref{sec:appendix-synthemol}.


\subsection{Guided Monte Carlo Tree Search}
The pioneering work that integrates neural networks with MCTS is AlphaGo \citep{AlphaGo}. Building upon that, AlphaZero refined this approach by merging the policy and value networks into a single, more efficient neural network \citep{AlphaZero}. What's more, AlphaZero was trained exclusively through self-play, enabling the system to adapt and optimize its game play without relying on external data. AlphaZero provides a highly relevant framework for the application to molecular generation. Although our approach diverges from AlphaZero in that it does not integrate a value network within the MCTS framework—instead relying on separate property predictors—the role of the policy network remains pivotal in both contexts. In molecular generation, our policy network is instrumental in selecting the next building block for synthesis, which directly influences the structure and properties of the resultant molecule, akin to how each move in Go influences the progression and outcome of the game.

\subsection{Retrosynthesis Evaluation}
While the ideal method to validate the synthesis pathway of our generated products would be through experimental testing in a laboratory, practical constraints currently prevent us from doing so. Consequently, we rely on in silico evaluations to estimate synthesizability. We first calculate the synthetic accessibility score (SA score) to predict synthesizability in an automated manner \citep{SA-score}. Additionally, we seek to validate our proposed synthesis pathways using Syntheseus \citep{syntheseus}, a python package for retrosynthetic planning. 

The SA score is calculated through a combination of fragment contributions and a complexity penalty \citep{SA-score}. The resulting SA score provides a metric for synthetic accessibility, ranging from 1 (indicating easy synthesis) to 10 (indicating high difficulty). However, while the SA score differentiates between feasible and infeasible molecules to some extent, it does not provide specific insights into the actual synthesis pathways. Syntheseus operates by recursively decomposing a target molecule into increasingly simpler molecules through a backward reaction prediction model, continuing until it identifies a set of synthetically accessible building blocks \citep{syntheseus}. In our experiments, we employ the Molecule Edit Graph Attention Network (MEGAN) backward reaction prediction model \citep{MEGAN}.

\section{Methodology}
This section presents the lipid building block dataset construction, the lipid property predictors, the reaction prediction model, and the policy network guided MCTS approach for lipid generation.

\subsection{Dataset Construction}
\label{sec:dataset-construction}
An ionizable lipid consists of an ionizable lipid head and several lipid tails. Our building block datasets contain synthetically accessible molecules that can act as lipid heads or lipid tails. We filter valid lipid building blocks from the ZINC20 database \citep{ZINC20}, a large-scale chemical database designed for drug discovery and virtual screening.

\begin{figure}
    \centering
    \includegraphics[width=0.9\linewidth]{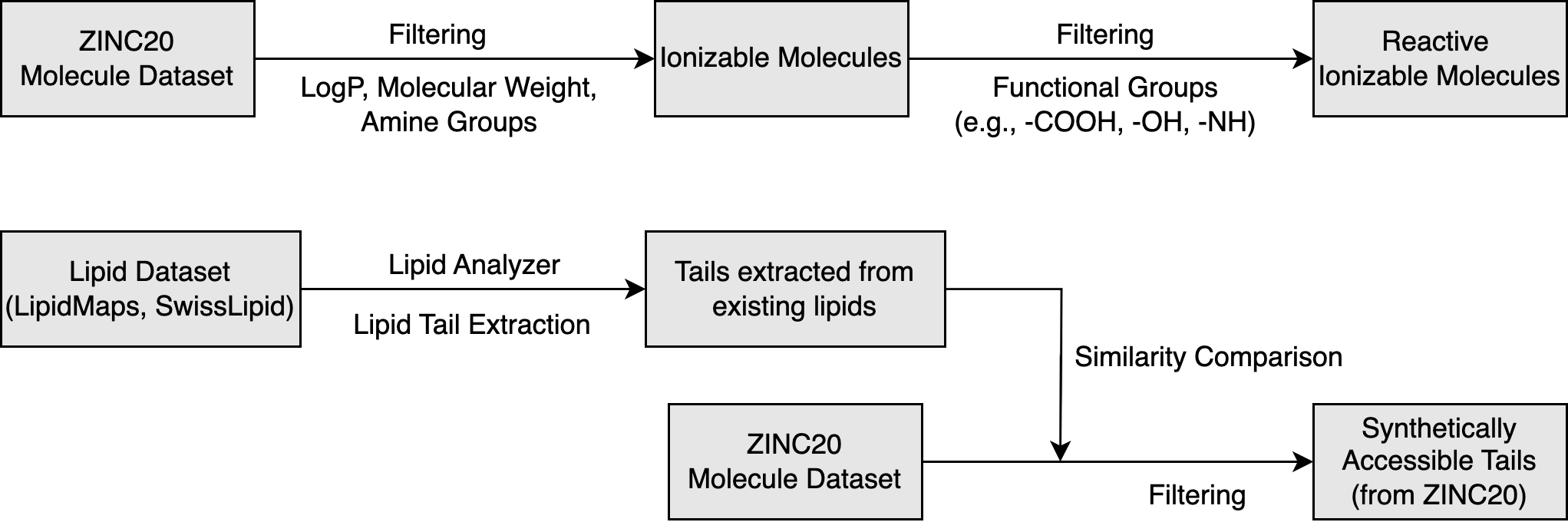}
    \caption{Roadmap of lipid building block datasets construction. The upper flowchart illustrates the process of filtering ionizable lipid head building block dataset. The lower flowchart illustrates the process of constructing lipid tail building block dataset.}
    \label{fig:RoadMap-DatasetConstruction}
\end{figure}

Figure \ref{fig:RoadMap-DatasetConstruction} shows the flowcharts for constructing lipid head and lipid tail building block datasets. For the ionizable lipid head dataset, we first filter ionizable molecules from the ZINC20 molecule dataset based on specific criteria of LogP value, molecular weight, and amine groups. Since the lipid head building blocks are expected to react with lipid tail building blocks, we further filter reactive molecules by identifying whether or not the molecule contains certain functional groups that participate in common reactions. Detailed filtering criteria are presented in Appendix \ref{sec:appendix-dataset-example}.

We identify a lipid tail building block if the molecule is similar to an actual lipid tail (i.e., the tail component of an existing lipid). Taking advantage of the Lipid Analyzer\footnote{The Lipid Analyzer toolkit is an unpublished work.}, an implemented toolkit for lipid tail extraction, we extract lipid tails from a lipid dataset sourced from \citep{LipidMaps, SwissLipids}. We then conduct a similarity comparison to filter those molecules from the ZINC20 dataset that are similar to a real lipid tail. The resulting molecules make up our lipid tail building block dataset. Detailed information of the dataset construction process and selective examples of lipid building blocks are shown in Appendix \ref{sec:appendix-dataset-example}.

\subsection{Lipid Property Prediction}
Our generative approach relies on molecular property predictors to evaluate the potential of generated molecules to be an ionizable lipid. We employ a lipid classifier for binary assessment of lipid-likeness and an ionizability predictor to evaluate whether the generated product is ionizable. 

\textbf{Lipid Classifier} Utilizing the Chemprop graph neural network framework \citep{Chemprop}, our lipid classifier features three message passing layers that integrate molecular features, complemented by two feed-forward layers dedicated to predicting molecular properties. The dataset we use to train the lipid classifier contains 180\,000 lipid samples and 180\,000 non-lipid samples. The non-lipid samples are small molecules from the PubChem database \citep{PubChem}. Part of the lipid samples come from publicly accessible lipid dataset LipidMaps and SwissLipids \citep{LipidMaps, SwissLipids}. The rest of the lipid samples are generated using a hierarchical graph encoder-decoder that employs significantly large and flexible graph motifs as basic building blocks \citep{AIGenLipidStrategy}.

\textbf{Ionizability Predictor} An ionizable molecule has neutral charge in physiological pH and becomes positively charged in acidic environments \citep{carrasco_ionization_2021, stern_nanotechnology_safety_2007}. For simpler filtering purpose, we only consider the net charge under pH = 7.4 (i.e., represents the physiological pH) and pH = 5 (i.e., represents the acidic environment). We utilize the MolGpKa module to determine acidic and basic groups within a given molecule and to estimate their respective pKa values \citep{MolGpKa}. The net charge of each molecule under a given pH value can then be calculated using the Henderson-Hasselbalch equation \citep{GeneralChemistry-HHeq}. 

\subsection{Reaction Prediction}
We employ a template-based reaction prediction model which applies predefined reaction templates—derived from known chemical reaction mechanisms—to reactants. The advantages of this approach include faster computation, adherence to established chemical rules, and higher synthesis success potential. We adopt the same reaction templates as SyntheMol, our baseline model. The combinatorial chemical space explored by SyntheMol was the Enamine REadily AccessibLe (REAL) Space \citep{Enamine-Real-space}. SyntheMol employs 13 reactions that account for 93.9\% of the REAL Space \citep{SyntheMol}. Although our lipid building block dataset differs from the REAL space, the selected reactions are widely applicable across a broad range of chemicals, making them suitable for our project as well.

\subsection{Guided Monte Carlo Tree Search for Lipid Generation}
Building upon the SyntheMol and inspired by AlphaZero \citep{SyntheMol, AlphaZero}, we apply the policy network guided MCTS to ionizable lipid generation. The integration of a policy network with MCTS forms the cornerstone of our approach, enabling a strategic exploration of chemical spaces through guided decision-making. In our problem formulation, the state is defined as the current molecule, and the action as the selection of the next building block. The policy network assigns probabilities to potential actions for each state. The training of the policy network is a cyclic process, aimed at progressively improving the network’s ability to predict and prioritize effective synthetic pathways.

\begin{figure}
    \centering
    \includegraphics[width=0.6\linewidth]{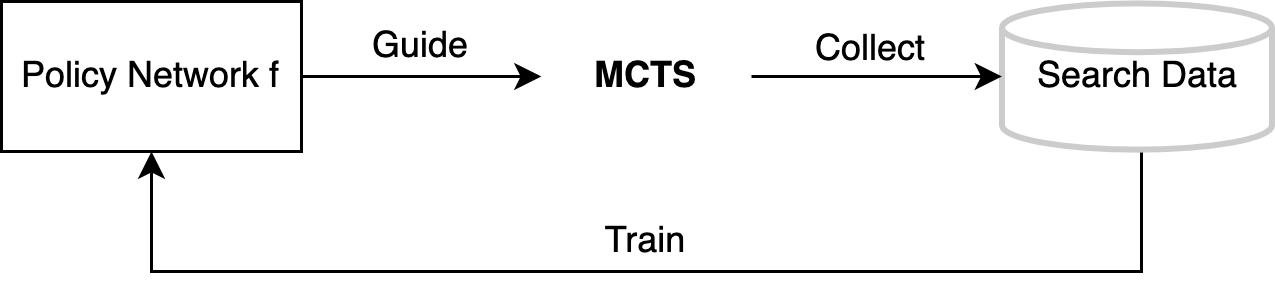}
    \caption{Workflow of policy network training.}
    \label{fig:workflow-policy-network-training}
\end{figure}

Figure \ref{fig:workflow-policy-network-training} illustrates the workflow of the policy network training procedure. We start from a randomly initialized policy network which assigns equal probabilities to all the actions in the provided action space. This policy network will be used to guide the MCTS. We conduct the tree search for a number of simulations, and the search data (i.e., visit counts of all state-action pairs involved) of the tree search will be used as the training data to train the policy network for several epochs. We propose a customized policy network training technique, which is detailed in Appendix \ref{chapter:appendix-policy-network-training}. This process serves as one iteration of policy network training. In the next iteration, we use the updated policy network to guide MCTS and repeat the process.

A step-wise depiction of each simulation of the MCTS is shown in Figure \ref{fig:guided-MCTS}, with the detailed algorithm presented in Appendix \ref{sec:appendix-algo-guided-mcts}. Each simulation of the MCTS consists of four steps: select, expand, rollout, and backpropagate. We define the policy network that guides the MCTS simulations to be $f_\theta$ with parameters $\theta$. Each edge in the tree search represents a state-action pair $(s, a)$ where state $s$ is the current molecule we have and action $a$ is the next building block molecule to choose. Each edge stores a set of statistics $\{N(s, a), W(s, a), P(s, a)\}$ where $N(s, a)$ is the visit count, $W(s, a)$ is the total action value (i.e., sum of values of final products reached after taking action $a$ from state $s$), and $P(s, a)$ is the prior probability of selecting that edge. Note that $P(s, \cdot) = f_\theta (s)$ is given by the current policy network.

\begin{figure}
    \centering
    \includegraphics[width=0.9\linewidth]{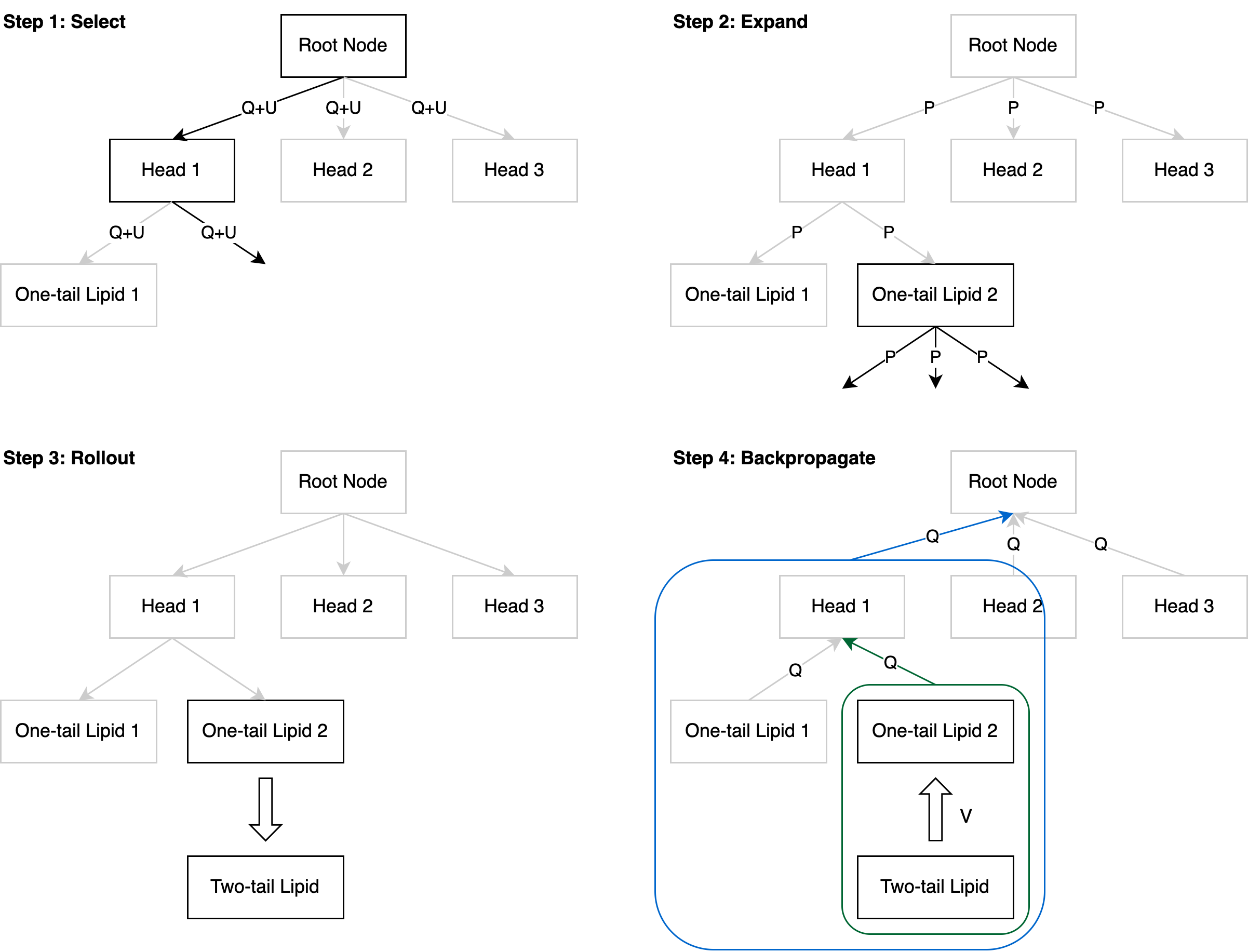}
    \caption{Policy network guided Monte Carlo tree search for lipid generation.}
    \label{fig:guided-MCTS}
\end{figure}

In the selection step, each simulation traverses the tree by selecting the edge with the maximum Upper Confidence Bound (UCB) score until a leaf node is reached. At each time step $t$, our action selection criterion follows
\begin{equation}
    a_t = \argmax_a UCB\_score(s_t, a) = \argmax_a (Q(s_t, a) + U(s_t, a))
\end{equation}
The UCB score is defined to be
\begin{align}
\label{eq:guided-mcts-ucb}
    UCB\_score(s, a) &= Q(s, a) + U(s, a) \\
                     &= \frac{W(s, a)}{N(s, a)} + c \cdot P(s, a) \cdot\frac{\sqrt{\sum_b N(s, b)}}{1 + N(s, a)}
\end{align}
where $c$ is a constant parameter controlling the level of exploration. 

In the expansion step, the leaf node selected in the previous step will be expanded. First, the state $s$ of the leaf node will be calculated as the chemical product of the state and action molecules represented by the edge which directs to this leaf node. We then find the action space $A(s)$ of the leaf node and calculate $P(s, a)$ values for all $a \in A(s)$ via the policy network, i.e., $P(s, \cdot) = f_\theta (s)$. The $P$ values will be stored in the newly-added outgoing edges from the selected leaf node.

Meanwhile, the selected leaf node will be evaluated by the rollout step. This step aims to get a value for the selected leaf node. The rollout means performing random actions until we reach the end of the play (i.e., until we generate a two-tail lipid). This randomly generated product will be evaluated by the property predictor and this property score will act as the value of the selected leaf node.

Once we have the value of the selected leaf node, we perform the last step of the simulation, backpropagation. The value will be backpropagated along the chosen path to update action values $Q$.

\section{Experiments}
Our experimental investigation focuses on comparing the performance of the SyntheMol approach, served as the baseline, with our proposed policy network guided MCTS approach. We demonstrate the enhanced efficiency of the guided MCTS approach in identifying and synthesizing high-potential ionizable lipids.

\subsection{Experimental Setups}
\textbf{Lipid Building Block Dataset} Following the procedure described in Section \ref{sec:dataset-construction}, we construct a lipid building block dataset consisting of over 2.7 million lipid head building blocks and 5310 lipid tail building blocks. While directly incorporating such huge dataset into the MCTS would result in a predominantly exploratory behavior akin to random selections, we extract a subset of approximately 12\,000 head building blocks to define our actual head search space. The entire tail building block dataset is utilized in subsequent levels of the search. For the evaluation of the policy network in the guided MCTS approach, an additional set of 200 testing head building blocks is employed.

\textbf{SyntheMol Configuration}
We constrain the maximum number of child node expansions to be 2\,000, meaning that each MCTS explores a head search space of this size. The MCTS is executed over 10\,000 simulations to ensure comprehensive analysis of the generated lipid products. Additionally, the exploration weight $c$ used in the UCB score calculation, as detailed in Equation \ref{eq:naive-mcts-ucb}, is set at 10. 

\textbf{Guided MCTS Configuration}
We operate the guided MCTS across 10 iterations. For every iteration, the MCTS is executed 10 times, each exploring a head space of 200. This setup allows the 10 runs of MCTS collectively to explore a total head search space size of 2\,000, aligning with the head search space used in the SyntheMol approach. Each MCTS runs 10\,000 simulations to gather substantial search data. The data from these 10 MCTS runs are pooled to train the policy network, and the accumulated generated products are analyzed as the iteration's output. The policy network undergoes 20 epochs of training in each iteration. The exploration weight $c$ used in the UCB score calculation, as outlined in Equation \ref{eq:guided-mcts-ucb}, is set at 20.

\textbf{Policy Network Configuration}
The policy network processes state-action pairs, where each state is the current molecule and the action is the next building block molecule selected. Each molecule is represented using a Morgan fingerprint with 1024-bit binary digit and radius 2 \citep{fingerprint}, and the fingerprints of both the state and action molecules are concatenated to form a 2048-bit feature vector, serving as the input to the policy network. Our policy network architecture consists of four linear layers, each followed by a ReLU activation function \citep{Relu} and dropout layers with a dropout rate of 0.5 \citep{dropout} to prevent overfitting. The Adam optimizer is used for network training, with a learning rate of 0.001 \citep{Adam}.

\textbf{Computing Resources}
Our computational setup includes a GPU server equipped with 8 Tesla P-100 GPUs, each featuring 16 GB of memory. It is important to note that within our experimental framework, only the MolGpKa module is configured to utilize GPU resources \citep{MolGpKa}. All other components of our algorithms are designed to run efficiently on CPU.

\subsection{Results and Discussions}
We first present the performance of our lipid property predictors and then compare the generative results of the SyntheMol approach with our proposed policy network guided MCTS approach.

\subsubsection{Lipid Property Prediction}
As previously mentioned, we utilize a Chemprop based binary lipid classifier to  determine whether a candidate molecule is lipid-like \citep{Chemprop}. After training for a single epoch, our model demonstrates exceptional performance, achieving both Receiver Operating Characteristic Area Under the Curve (ROC-AUC) and Precision-Recall Area Under the Curve (PR-AUC) scores above 0.9999. Together, these metrics, along with a test accuracy of 99.89\%, underscore the robustness and predictive accuracy of our lipid classifier.

In terms of ionizability, we utilize the MolGpKa module to identify ionizable groups within a target molecule and calculates their corresponding pKa values. We then determine the net charge of the target molecule at specific pH levels and conduct the ionizability filtering. According to \citep{MolGpKa}, the MolGpKa predictor has undergone rigorous testing, thus justifying the credibility of our ionizability predictions.

To better validate our lipid property predictors, we curated an ionizable lipid dataset with more than 2\,500 ionizable lipids that have been experimentally synthesized from published works \citep{AGILE, published-data-1, published-data-2, published-data-3, published-data-4, published-data-5, published-data-6}. We evaluated our predictors on this dataset, achieving an accuracy of 98.32\% for our lipid classifier. Notably, all ionizable lipids were correctly classified by the ionizability predictor.

\subsubsection{Generative Results}
Our primary objective is to generate products that are potent candidates for ionizable lipids, meaning we aim for molecules with high property scores. We begin by examining the products generated via the SyntheMol approach, which serves as a baseline for comparing the performance enhancements achieved with the guided MCTS approach. 

The property score combines a lipid classifier score, which typically hovers near 0 or 1, and a binary ionizability score. A property score approaching 2 typically identifies the molecule as a likely ionizable lipid. After conducting 10\,000 simulations, a total of 16\,477 two-tail lipids were generated. We observe that only 4513 of the 16\,477 generated products are predicted to be ionizable lipids, providing an ionizable lipid rate at 0.2739. 
To evaluate the effectiveness of MCTS, we conducted experiments with random generation. Specifically, we generated 10\,000 unique two-tail lipids through random combinations using the same lipid head and tail building block subsets, achieving an ionizable lipid rate of 0.1547. This result justifies the effectiveness of MCTS in guiding the generation of ionizable lipids.

We now turn our attention to the outcomes of the policy network guided MCTS approach. By comparing these results with those from the SyntheMol approach, we seek to illustrate the enhanced effectiveness of the policy network guided MCTS in producing ionizable lipids. We analyze the unique generated products from both training and testing simulations. The primary distinction between these simulations lies in their respective head building block search spaces. However, it is important to note that the selection of head molecules significantly influences the quality of the generated products. Nonetheless, observing the trend in quality changes within a specific head set, as guided by different iterations of the policy network, can still effectively illustrate the efficiency and impact of policy network training.

\begin{figure}[ht]
    \centering
    \begin{subfigure}[b]{0.495\textwidth}
        \centering
        \includegraphics[width=\textwidth]{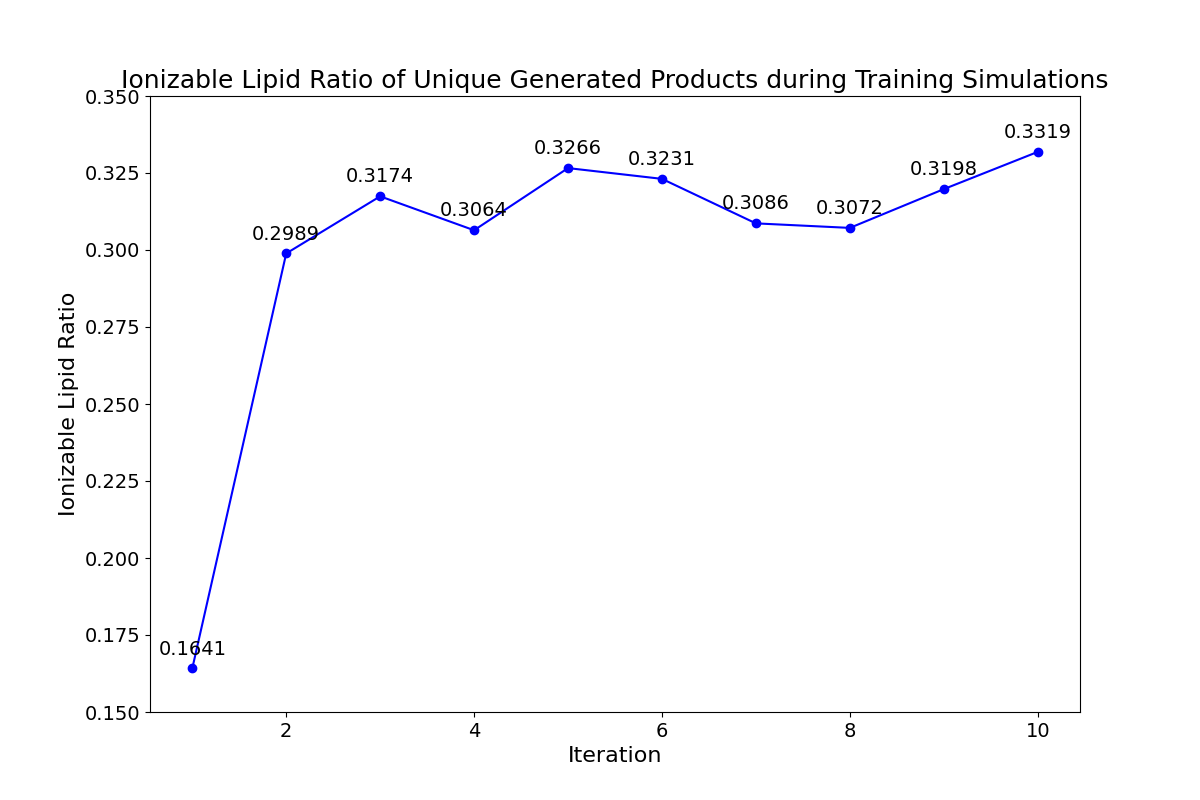}
        \caption{Ionizable lipid rate during training simulations}
        \label{fig:train_generation_ratio_unique}
    \end{subfigure}
    \begin{subfigure}[b]{0.495\textwidth}
        \centering
        \includegraphics[width=\textwidth]{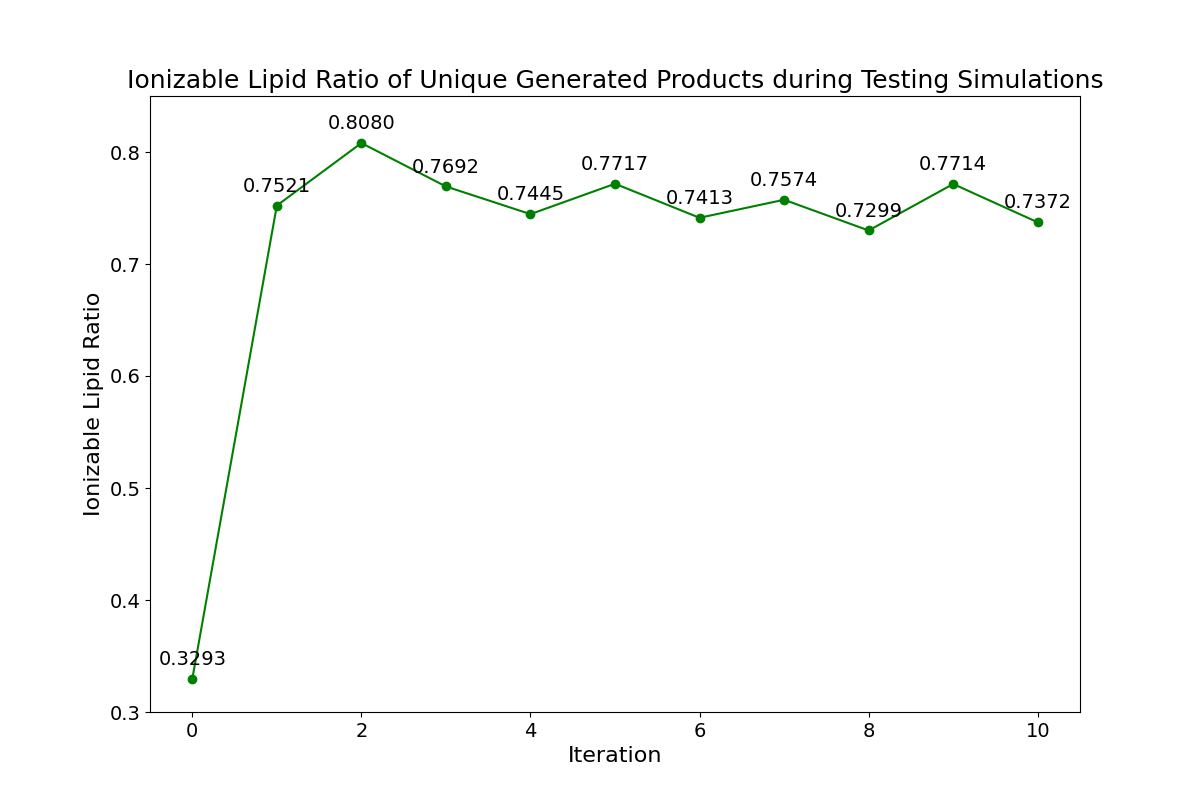}
        \caption{Ionizable lipid rate during testing simulations}
        \label{fig:test_generation_ratio_unique}
    \end{subfigure}
    \caption{Ionizable lipid rate of the generated products during guided MCTS simulations.}
    \label{fig:guided_mcts_generation}
\end{figure}

Figure \ref{fig:train_generation_ratio_unique} depicts the ionizable lipid rate cross training iterations. The initial results in iteration 1 are derived from MCTS simulations guided by a randomly initialized policy network, whereas the results from iteration 2 onward are influenced by successively trained instances of the policy network. A distinct increase in the ionizable lipid generation rate is evident after the first training session, as is shown in Figure \ref{fig:train_generation_ratio_unique}, which underscores a significant enhancement in quality attributable to the training of the policy network, affirming its efficacy in refining the generation process. Figure \ref{fig:test_generation_ratio_unique} presents the ionizable lipid rate during testing simulations, each guided by policy networks trained across different iterations. The initial iteration (i.e., iteration 0) features products generated by MCTS simulations directed by a randomly initialized policy network, with subsequent iterations using progressively trained policy networks. The results demonstrate a marked improvement in the ionizable lipid rate following the initial training iteration. Subsequent iterations show the rate stabilizing, with only minor fluctuations observed, oscillating between 0.73 and 0.8.  In either case, these rates significantly surpass the ionizable lipid rate of below 0.3 achieved by the SyntheMol approach, clearly demonstrating the superior performance of our model over the baseline. A detailed comparison of the ionizable lipid generation rate is presented in Table \ref{table:evaluation}, where the rate for the guided MCTS corresponds to the value recorded after 10 iterations of policy network training.

\subsubsection{Retrosynthesis Evaluation}

Table \ref{table:evaluation} presents the evaluation results of our generated ionizable lipids and their corresponding synthesis pathways. We consider all unique products generated during both training and testing phases across all iterations of the guided MCTS approach. For better comparison, we also include the generated products from random combinations and the dataset of published ionizable lipids  synthesized in previous works \citep{AGILE, Ionizable-lipid-source-1, Ionizable-lipid-source-2}.

The SA score, which ranges from 1 to 10 with lower values indicating easier synthesis \citep{SA-score}, reveals similar results for both the SyntheMol and guided MCTS approaches. Notably, there is no significant difference between the SA scores of our generated products and those of the published ionizable lipids. While the SA score justifies the synthesizability of the generate products to some extent, it does not offer any insights into their specific synthesis pathways.


Using Syntheseus, we aim to identify synthesis pathways for the generated products using a backward reaction prediction model. Given that a single generated product may have multiple possible synthesis pathways, we do not strictly adhere to those suggested by the generative model. Instead, we focus on identifying reactants for the generated products and checking whether these reactants are present in our lipid building block dataset. We record the proportion of generated products that can be synthesized using molecules from the building block dataset.

\begin{table}
\centering
\caption{Evaluation of generated products and synthesis pathways.}
\label{table:evaluation}
\begin{tabular}{ccccc}
\toprule
 & \begin{tabular}[c]{@{}c@{}}No. Unique\\ Ionizable Lipids \end{tabular} & \begin{tabular}[c]{@{}c@{}}Unique Ionizable\\ Lipid Rate \end{tabular} & \begin{tabular}[c]{@{}c@{}}Average\\ SA Score\end{tabular} & \begin{tabular}[c]{@{}c@{}}Retro Valid Rate\\ by Syntheseus\end{tabular}  \\ \midrule
\begin{tabular}[c]{@{}c@{}}Guided MCTS Train\end{tabular} & 5058 & 0.3319 & 4.62 & 0.4881  \\ 
\begin{tabular}[c]{@{}c@{}}Guided MCTS Test\end{tabular} & 545 & 0.7372 & 4.24 & 0.2679  \\ 
SyntheMol & 4513 & 0.2739 & 4.77 & 0.2719  \\ 
Random Combination & 1547 & 0.1547 & 4.44 & 0.3103 \\ 
Published Lipids & 2563 & / & 4.12 & 0.0433 \\ 
\bottomrule
\end{tabular}
\end{table}

We observe that the retrosynthesis rate validated by Syntheseus is not high, with none of the cases exceeding 50\%. The low rate may result from the unique characteristics of the lipid building block dataset and the reaction templates we adopted. For retrosynthesis planning, we utilize the MEGAN backward reaction prediction model trained on the USPTO-50k dataset \citep{MEGAN}. Our lipid building block molecules may differ significantly from the reactants in the USPTO reactions, and our reaction templates may not align well with those in the USPTO dataset. Additionally, a lot of the reaction templates involve using linker molecules, while such reactions might not be present in the USPTO dataset. These mismatches make retrosynthesis planning challenging and contribute to the low retrosynthesis rate. Regarding the published ionizable lipids that have been experimentally synthesized, we observe a retrosynthesis valid rate below 5\%. This is likely because these lipids were synthesized using building blocks outside our building block dataset. Our building block dataset may lack many structures not included in the ZINC dataset. Additionally, our choice of weight cutoff and restrictions on amine-containing structures during lipid head dataset construction further reduce the search space.

Selected examples of generated ionizable lipids, along with their synthesis pathways validated by the retrosynthesis tool, are presented in Appendix \ref{chapter:appendix-synthesis-example}.

\section{Conclusion}
In this work, we have explored the Monte Carlo tree search (MCTS)-based generative models for ionizable lipid generation. We constructed a lipid building block dataset, featuring synthetically accessible ionizable lipid heads and tails, which is well-suited for future lipid generation tasks. We developed two lipid property predictors: a lipid classifier and an ionizability predictor, both designed to accurately assess whether a candidate molecule is lipid-like or ionizable. Adapting the SyntheMol approach, originally utilized for antibiotic discovery, we tailored this method for lipid generation to serve as our baseline model. We further innovated by developing a policy network guided MCTS-based generative model which is capable of producing high-quality ionizable lipids with available synthesis paths, outperforming our baseline. Our achievements indicate that this project offers a promising direction for ionizable lipid generation, also contributing to the broader field of drug delivery.

Nonetheless, it is important to acknowledge that the work has its limitations. If condition allowed, we may conduct experimental validations of the existing reaction templates with lipid building blocks to ensure their applicability and efficiency in lipid synthesis. It will be helpful to develop and integrate additional reaction templates that are specific to lipid chemistry. To address the limitation of not having practical synthesis validations, future research should prioritize establishing collaborations with chemical laboratories. This will enable empirical testing and validation of the synthesized molecules, providing a direct assessment of their practical viability and safety. Develop or customize computer-aided retrosynthesis tools specifically for ionizable lipid generation can also be helpful.

\begin{ack}
The authors would like to thank Asal Mehradfar and Mohammad Shahab Sepehri for curating the lipid dataset used for training lipid classifier and for building the Lipid Analyzer toolkit.

\end{ack}

\bibliographystyle{plainnat}
\bibliography{reference}

\begin{thebibliography}{55}
\providecommand{\natexlab}[1]{#1}
\providecommand{\url}[1]{\texttt{#1}}
\expandafter\ifx\csname urlstyle\endcsname\relax
  \providecommand{\doi}[1]{doi: #1}\else
  \providecommand{\doi}{doi: \begingroup \urlstyle{rm}\Url}\fi

\bibitem[Abd~Elwakil et~al.(2023)Abd~Elwakil, Suzuki, Khalifa, Elshami, Isono, Elewa, Sato, Nakamura, Satoh, and Harashima]{published-data-4}
Mahmoud~M. Abd~Elwakil, Ryota Suzuki, Alaa~M. Khalifa, Rania~M. Elshami, Takuya Isono, Yaser~H.A. Elewa, Yusuke Sato, Takashi Nakamura, Toshifumi Satoh, and Hideyoshi Harashima.
\newblock Harnessing topology and stereochemistry of glycidylamine-derived lipid nanoparticles for in vivo mrna delivery to immune cells in spleen and their application for cancer vaccination.
\newblock \emph{Advanced Functional Materials}, 33\penalty0 (45):\penalty0 2303795, 2023.
\newblock \doi{https://doi.org/10.1002/adfm.202303795}.
\newblock URL \url{https://onlinelibrary.wiley.com/doi/abs/10.1002/adfm.202303795}.

\bibitem[Aimo et~al.(2015)Aimo, Liechti, Hyka-Nouspikel, Niknejad, Gleizes, Götz, Kuznetsov, David, van~der Goot, Riezman, Bougueleret, Xenarios, and Bridge]{SwissLipids}
Lucila Aimo, Robin Liechti, Nevila Hyka-Nouspikel, Anne Niknejad, Anne Gleizes, Lou Götz, Dmitry Kuznetsov, Fabrice P.~A. David, F.~Gisou van~der Goot, Howard Riezman, Lydie Bougueleret, Ioannis Xenarios, and Alan Bridge.
\newblock The swisslipids knowledgebase for lipid biology.
\newblock \emph{Bioinformatics}, 31\penalty0 (17):\penalty0 2860--2866, 2015.
\newblock \doi{10.1093/bioinformatics/btv285}.

\bibitem[Assouel et~al.(2018)Assouel, Ahmed, Segler, Saffari, and Bengio]{GAN-DEFactor}
Rim Assouel, Mohamed Ahmed, Marwin H.~S. Segler, Amir Saffari, and Yoshua Bengio.
\newblock Defactor: Differentiable edge factorization-based probabilistic graph generation.
\newblock \emph{CoRR}, abs/1811.09766, 2018.
\newblock URL \url{http://arxiv.org/abs/1811.09766}.

\bibitem[Bian et~al.(2019)Bian, Wang, Jun, and Xie]{dcGAN}
Yuemin Bian, Junmei Wang, Jaden~Jungho Jun, and Xiang-Qun Xie.
\newblock Deep convolutional generative adversarial network ({dcGAN}) models for screening and design of small molecules targeting cannabinoid receptors.
\newblock \emph{Molecular Pharmaceutics}, 16\penalty0 (11):\penalty0 4451--4460, 2019.
\newblock ISSN 1543-8384.
\newblock \doi{10.1021/acs.molpharmaceut.9b00500}.
\newblock URL \url{https://doi.org/10.1021/acs.molpharmaceut.9b00500}.
\newblock Publisher: American Chemical Society.

\bibitem[Bradshaw et~al.(2020)Bradshaw, Paige, Kusner, Segler, and Hernández-Lobato]{DOG}
John Bradshaw, Brooks Paige, Matt~J. Kusner, Marwin H.~S. Segler, and José~Miguel Hernández-Lobato.
\newblock Barking up the right tree: an approach to search over molecule synthesis dags, 2020.

\bibitem[Carrasco et~al.(2021)Carrasco, Alishetty, Alameh, Said, Wright, Paige, Soliman, Weissman, Cleveland, Grishaev, and Buschmann]{carrasco_ionization_2021}
Manuel~J. Carrasco, Suman Alishetty, Mohamad-Gabriel Alameh, Hooda Said, Lacey Wright, Mikell Paige, Ousamah Soliman, Drew Weissman, Thomas~E. Cleveland, Alexander Grishaev, and Michael~D. Buschmann.
\newblock Ionization and structural properties of {mRNA} lipid nanoparticles influence expression in intramuscular and intravascular administration.
\newblock \emph{Communications Biology}, 4\penalty0 (1):\penalty0 956, 2021.
\newblock ISSN 2399-3642.
\newblock \doi{10.1038/s42003-021-02441-2}.
\newblock URL \url{https://doi.org/10.1038/s42003-021-02441-2}.

\bibitem[Coulom(2006)]{Coulom2006EfficientSA}
R{\'e}mi Coulom.
\newblock Efficient selectivity and backup operators in monte-carlo tree search.
\newblock In \emph{Computers and Games}, 2006.
\newblock URL \url{https://api.semanticscholar.org/CorpusID:16724115}.

\bibitem[Degors et~al.(2019)Degors, Wang, Rehman, and Zuhorn]{ionizable-lipid-2}
Isabelle M.~S. Degors, Cuifeng Wang, Zia~Ur Rehman, and Inge~S. Zuhorn.
\newblock Carriers break barriers in drug delivery: Endocytosis and endosomal escape of gene delivery vectors.
\newblock \emph{Accounts of Chemical Research}, 52\penalty0 (7):\penalty0 1750--1760, 2019.
\newblock ISSN 0001-4842.
\newblock \doi{10.1021/acs.accounts.9b00177}.
\newblock URL \url{https://doi.org/10.1021/acs.accounts.9b00177}.
\newblock Publisher: American Chemical Society.

\bibitem[Ding et~al.(2023)Ding, Zhang, Jia, and Sun]{ding2023-TE}
Daisy~Yi Ding, Yuhui Zhang, Yuan Jia, and Jiuzhi Sun.
\newblock Machine learning-guided lipid nanoparticle design for mrna delivery, 2023.
\newblock URL \url{https://arxiv.org/abs/2308.01402}.

\bibitem[Ertl and Schuffenhauer(2009)]{SA-score}
Peter Ertl and Ansgar Schuffenhauer.
\newblock Estimation of synthetic accessibility score of drug-like molecules based on molecular complexity and fragment contributions.
\newblock \emph{Journal of Cheminformatics}, 1\penalty0 (1):\penalty0 8, 2009.
\newblock ISSN 1758-2946.
\newblock \doi{10.1186/1758-2946-1-8}.
\newblock URL \url{https://doi.org/10.1186/1758-2946-1-8}.

\bibitem[Fukushima(1969)]{Relu}
Kunihiko Fukushima.
\newblock Visual feature extraction by a multilayered network of analog threshold elements.
\newblock \emph{IEEE Trans. Syst. Sci. Cybern.}, 5:\penalty0 322--333, 1969.
\newblock URL \url{https://api.semanticscholar.org/CorpusID:206799280}.

\bibitem[Gao and Coley(2020)]{gao2020synthesizability}
Wenhao Gao and Connor~W. Coley.
\newblock The synthesizability of molecules proposed by generative models, 2020.
\newblock URL \url{https://arxiv.org/abs/2002.07007}.

\bibitem[Grygorenko et~al.(2020)Grygorenko, Radchenko, Dziuba, Chuprina, Gubina, and Moroz]{Enamine-Real-space}
Oleksandr~O. Grygorenko, Dmytro~S. Radchenko, Igor Dziuba, Alexander Chuprina, Kateryna~E. Gubina, and Yurii~S. Moroz.
\newblock Generating multibillion chemical space of readily accessible screening compounds.
\newblock \emph{iScience}, 23\penalty0 (11):\penalty0 101681, 2020.
\newblock \doi{10.1016/j.isci.2020.101681}.
\newblock Published correction appears in iScience. 2020 Dec 04;23(12):101873. doi: 10.1016/j.isci.2020.101873.

\bibitem[Gómez-Bombarelli et~al.(2018)Gómez-Bombarelli, Wei, Duvenaud, Hernández-Lobato, Sánchez-Lengeling, Sheberla, Aguilera-Iparraguirre, Hirzel, Adams, and Aspuru-Guzik]{CVAE}
Rafael Gómez-Bombarelli, Jennifer~N. Wei, David Duvenaud, José~Miguel Hernández-Lobato, Benjamín Sánchez-Lengeling, Dennis Sheberla, Jorge Aguilera-Iparraguirre, Timothy~D. Hirzel, Ryan~P. Adams, and Alán Aspuru-Guzik.
\newblock Automatic chemical design using a data-driven continuous representation of molecules.
\newblock \emph{ACS Central Science}, 4\penalty0 (2):\penalty0 268–276, January 2018.
\newblock ISSN 2374-7951.
\newblock \doi{10.1021/acscentsci.7b00572}.
\newblock URL \url{http://dx.doi.org/10.1021/acscentsci.7b00572}.

\bibitem[He et~al.(2023)He, Le, Shi, Liu, Liu, and Chen]{published-data-2}
Zepeng He, Zhicheng Le, Yi~Shi, Lixin Liu, Zhijia Liu, and Yongming Chen.
\newblock A multidimensional approach to modulating ionizable lipids for high-performing and organ-selective mrna delivery.
\newblock \emph{Angewandte Chemie International Edition}, 62\penalty0 (43):\penalty0 e202310401, 2023.
\newblock \doi{https://doi.org/10.1002/anie.202310401}.
\newblock URL \url{https://onlinelibrary.wiley.com/doi/abs/10.1002/anie.202310401}.

\bibitem[Huayamares et~al.(2023)Huayamares, Lokugamage, Rab, {Da Silva Sanchez}, Kim, Radmand, Loughrey, Lian, Hou, Achyut, Ehrhardt, Hong, Sago, Paunovska, Echeverri, Vanover, Santangelo, Sorscher, and Dahlman]{Ionizable-lipid-source-1}
Sebastian~G. Huayamares, Melissa~P. Lokugamage, Regina Rab, Alejandro~J. {Da Silva Sanchez}, Hyejin Kim, Afsane Radmand, David Loughrey, Liming Lian, Yuning Hou, Bhagelu~R. Achyut, Annette Ehrhardt, Jeong~S. Hong, Cory~D. Sago, Kalina Paunovska, Elisa~Schrader Echeverri, Daryll Vanover, Philip~J. Santangelo, Eric~J. Sorscher, and James~E. Dahlman.
\newblock High-throughput screens identify a lipid nanoparticle that preferentially delivers mrna to human tumors in vivo.
\newblock \emph{Journal of Controlled Release}, 357:\penalty0 394--403, 2023.
\newblock ISSN 0168-3659.
\newblock \doi{https://doi.org/10.1016/j.jconrel.2023.04.005}.
\newblock URL \url{https://www.sciencedirect.com/science/article/pii/S0168365923002559}.

\bibitem[Irwin et~al.(2020)Irwin, Tang, Young, Dandarchuluun, Wong, Khurelbaatar, Moroz, Mayfield, and Sayle]{ZINC20}
John~J. Irwin, Khanh~G. Tang, Jennifer Young, Chinzorig Dandarchuluun, Benjamin~R. Wong, Munkhzul Khurelbaatar, Yurii~S. Moroz, John Mayfield, and Roger~A. Sayle.
\newblock Z{I}{N}{C}20—a free ultralarge-scale chemical database for ligand discovery.
\newblock \emph{Journal of Chemical Information and Modeling}, 60\penalty0 (12):\penalty0 6065--6073, 2020.
\newblock \doi{10.1021/acs.jcim.0c00675}.
\newblock URL \url{https://doi.org/10.1021/acs.jcim.0c00675}.
\newblock PMID: 33118813.

\bibitem[Jin et~al.(2020)Jin, Barzilay, and Jaakkola]{AIGenLipidStrategy}
Wengong Jin, Regina Barzilay, and Tommi~S. Jaakkola.
\newblock Hierarchical generation of molecular graphs using structural motifs.
\newblock \emph{CoRR}, abs/2002.03230, 2020.
\newblock URL \url{https://arxiv.org/abs/2002.03230}.

\bibitem[Kim et~al.(2021)Kim, Jeong, Hur, Cho, Park, Jung, Seo, Woo, Nam, Lee, and Lee]{ionizable-lipid-1}
M.~Kim, M.~Jeong, S.~Hur, Y.~Cho, J.~Park, H.~Jung, Y.~Seo, H.~A. Woo, K.~T. Nam, K.~Lee, and H.~Lee.
\newblock Engineered ionizable lipid nanoparticles for targeted delivery of rna therapeutics into different types of cells in the liver.
\newblock \emph{Science Advances}, 7\penalty0 (9):\penalty0 eabf4398, 2021.
\newblock \doi{10.1126/sciadv.abf4398}.
\newblock URL \url{https://www.science.org/doi/abs/10.1126/sciadv.abf4398}.

\bibitem[Kim et~al.(2016)Kim, Thiessen, Bolton, Chen, Fu, Gindulyte, Han, He, He, Shoemaker, Wang, Yu, Zhang, and Bryant]{PubChem}
Sunghwan Kim, Paul~A. Thiessen, Evan~E. Bolton, Jie Chen, Gang Fu, Asta Gindulyte, Lianyi Han, Jane He, Siqian He, Benjamin~A. Shoemaker, Jiyao Wang, Bo~Yu, Jian Zhang, and Stephen~H. Bryant.
\newblock Pubchem substance and compound databases.
\newblock \emph{Nucleic Acids Research}, 44\penalty0 (D1):\penalty0 D1202--D1213, 2016.
\newblock \doi{10.1093/nar/gkv951}.

\bibitem[Kingma and Ba(2017)]{Adam}
Diederik~P. Kingma and Jimmy Ba.
\newblock Adam: A method for stochastic optimization, 2017.
\newblock URL \url{https://arxiv.org/abs/1412.6980}.

\bibitem[Kocsis and Szepesvari(2006)]{UCT}
Levente Kocsis and Csaba Szepesvari.
\newblock Bandit based monte-carlo planning.
\newblock In \emph{European Conference on Machine Learning}, 2006.
\newblock URL \url{https://api.semanticscholar.org/CorpusID:15184765}.

\bibitem[Kong et~al.(2023)Kong, Kim, Kim, Suk, and Yang]{mRNA-KONG2023114993}
Byoungjae Kong, Yelee Kim, Eun~Hye Kim, Jung~Soo Suk, and Yoosoo Yang.
\newblock mrna: A promising platform for cancer immunotherapy.
\newblock \emph{Advanced Drug Delivery Reviews}, 199:\penalty0 114993, 2023.
\newblock ISSN 0169-409X.
\newblock \doi{https://doi.org/10.1016/j.addr.2023.114993}.
\newblock URL \url{https://www.sciencedirect.com/science/article/pii/S0169409X23003083}.

\bibitem[Kularatne et~al.(2022)Kularatne, Crist, and Stern]{LNP}
Ruvanthi Kularatne, Rachael Crist, and Stephan Stern.
\newblock The future of tissue-targeted lipid nanoparticle-mediated nucleic acid delivery.
\newblock \emph{Pharmaceuticals}, 15:\penalty0 897, 07 2022.
\newblock \doi{10.3390/ph15070897}.

\bibitem[Kusner et~al.(2017)Kusner, Paige, and Hernández-Lobato]{GVAE}
Matt~J. Kusner, Brooks Paige, and José~Miguel Hernández-Lobato.
\newblock Grammar variational autoencoder, 2017.

\bibitem[Li et~al.(2023)Li, Manan, Liang, Gordon, Jiang, Varley, Gao, Langer, Xue, and Anderson]{published-data-1}
Bowen Li, Rajith~Singh Manan, Shun-Qing Liang, Akiva Gordon, Allen Jiang, Andrew Varley, Guangping Gao, Robert Langer, Wen Xue, and Daniel Anderson.
\newblock Combinatorial design of nanoparticles for pulmonary {mRNA} delivery and genome editing.
\newblock \emph{Nature Biotechnology}, 41\penalty0 (10):\penalty0 1410--1415, 2023.
\newblock ISSN 1546-1696.
\newblock \doi{10.1038/s41587-023-01679-x}.
\newblock URL \url{https://doi.org/10.1038/s41587-023-01679-x}.

\bibitem[Li et~al.(2024)Li, Raji, Gordon, Sun, Raimondo, Oladimeji, Jiang, Varley, Langer, and Anderson]{Bowen-TE}
Bowen Li, Idris Raji, Akiva Gordon, Lizhuang Sun, Theresa Raimondo, Favour Oladimeji, Allen Jiang, Andrew Varley, Robert Langer, and Daniel Anderson.
\newblock Accelerating ionizable lipid discovery for mrna delivery using machine learning and combinatorial chemistry.
\newblock \emph{Nature Materials}, 23:\penalty0 1--7, 05 2024.
\newblock \doi{10.1038/s41563-024-01867-3}.

\bibitem[Liu et~al.(2021)Liu, Cheng, Wei, Yu, Johnson, Farbiak, and Siegwart]{published-data-3}
Shuai Liu, Qiang Cheng, Tuo Wei, Xueliang Yu, Lindsay~T. Johnson, Lukas Farbiak, and Daniel~J. Siegwart.
\newblock Membrane-destabilizing ionizable phospholipids for organ-selective {mRNA} delivery and {CRISPR}–cas gene editing.
\newblock \emph{Nature Materials}, 20\penalty0 (5):\penalty0 701--710, 2021.
\newblock ISSN 1476-4660.
\newblock \doi{10.1038/s41563-020-00886-0}.
\newblock URL \url{https://doi.org/10.1038/s41563-020-00886-0}.

\bibitem[Luo et~al.(2021)Luo, Shi, Xu, and Tang]{Diffusion-2}
Shitong Luo, Chence Shi, Minkai Xu, and Jian Tang.
\newblock Predicting molecular conformation via dynamic graph score matching.
\newblock In M.~Ranzato, A.~Beygelzimer, Y.~Dauphin, P.S. Liang, and J.~Wortman Vaughan, editors, \emph{Advances in Neural Information Processing Systems}, volume~34, pages 19784--19795. Curran Associates, Inc., 2021.
\newblock URL \url{https://proceedings.neurips.cc/paper_files/paper/2021/file/a45a1d12ee0fb7f1f872ab91da18f899-Paper.pdf}.

\bibitem[Ly et~al.(2022)Ly, Daniel, Soriano, Kis, and Blakney]{Ionizable-lipid-source-2}
Han~Han Ly, Simon Daniel, Shekinah K.~V. Soriano, Zoltán Kis, and Anna~K. Blakney.
\newblock Optimization of lipid nanoparticles for sarna expression and cellular activation using a design-of-experiment approach.
\newblock \emph{Molecular Pharmaceutics}, 19\penalty0 (6):\penalty0 1892--1905, 2022.
\newblock \doi{10.1021/acs.molpharmaceut.2c00032}.
\newblock URL \url{https://doi.org/10.1021/acs.molpharmaceut.2c00032}.
\newblock PMID: 35604765.

\bibitem[Maziarz et~al.(2023)Maziarz, Tripp, Liu, Stanley, Xie, Gai{\'n}ski, Seidl, and Segler]{syntheseus}
Krzysztof Maziarz, Austin Tripp, Guoqing Liu, Megan Stanley, Shufang Xie, Piotr Gai{\'n}ski, Philipp Seidl, and Marwin Segler.
\newblock Re-evaluating retrosynthesis algorithms with syntheseus.
\newblock In \emph{NeurIPS 2023 AI for Science Workshop}, 2023.
\newblock URL \url{https://openreview.net/forum?id=W5U18rgtpg}.

\bibitem[Mitchell et~al.(2021)Mitchell, Billingsley, Haley, Wechsler, Peppas, and Langer]{nanoparticles-for-drug-delivery}
Michael~J. Mitchell, Margaret~M. Billingsley, Rebecca~M. Haley, Marissa~E. Wechsler, Nicholas~A. Peppas, and Robert Langer.
\newblock Engineering precision nanoparticles for drug delivery.
\newblock \emph{Nature Reviews Drug Discovery}, 20\penalty0 (2):\penalty0 101--124, 2021.
\newblock ISSN 1474-1784.
\newblock \doi{10.1038/s41573-020-0090-8}.
\newblock URL \url{https://doi.org/10.1038/s41573-020-0090-8}.

\bibitem[Moayedpour et~al.(2024)Moayedpour, Broadbent, Riahi, Bailey, Thu, Dobchev, Balsubramani, Santos, Kogler-Anele, Corrochano-Navarro, Li, Montoya, Agarwal, Bar-Joseph, and Jager]{TE-LLM}
Saeed Moayedpour, Jonathan Broadbent, Saleh Riahi, Michael Bailey, Hoa Thu, Dimitar Dobchev, Akshay Balsubramani, Ricardo Santos, Lorenzo Kogler-Anele, Alejandro Corrochano-Navarro, Sizhen Li, Fernando Montoya, Vikram Agarwal, Ziv Bar-Joseph, and Sven Jager.
\newblock Representations of lipid nanoparticles using large language models for transfection efficiency prediction.
\newblock \emph{Bioinformatics (Oxford, England)}, 40, 05 2024.
\newblock \doi{10.1093/bioinformatics/btae342}.

\bibitem[Pan et~al.(2021)Pan, Wang, Li, Zhang, and Ji]{MolGpKa}
Xiaolin Pan, Hao Wang, Cuiyu Li, John Zeng~Hui Zhang, and Changge Ji.
\newblock Mol{G}pka: A web server for small molecule pka prediction using a graph-convolutional neural network.
\newblock \emph{Journal of chemical information and modeling}, 2021.
\newblock URL \url{https://api.semanticscholar.org/CorpusID:235798026}.

\bibitem[Petrucci et~al.(2002)Petrucci, Harwood, and Herring]{GeneralChemistry-HHeq}
Ralph~H. Petrucci, William~S. Harwood, and F.~Geoffrey Herring.
\newblock \emph{General Chemistry: Principles and Modern Applications}.
\newblock Prentice Hall, Upper Saddle River, NJ, 8th edition, 2002.
\newblock ISBN 0-13-014329-4, 978-0-13-014329-7, 0-13-017677-X, 978-0-13-017677-6, 0-13-111673-8, 978-0-13-111673-3.
\newblock 1 volume (various pagings): illustrations (some color); 26 cm.

\bibitem[Sacha et~al.(2020)Sacha, Blaz, Byrski, Wlodarczyk{-}Pruszynski, and Jastrzebski]{MEGAN}
Mikolaj Sacha, Mikolaj Blaz, Piotr Byrski, Pawel Wlodarczyk{-}Pruszynski, and Stanislaw Jastrzebski.
\newblock Molecule edit graph attention network: Modeling chemical reactions as sequences of graph edits.
\newblock \emph{CoRR}, abs/2006.15426, 2020.
\newblock URL \url{https://arxiv.org/abs/2006.15426}.

\bibitem[Sahin et~al.(2014)Sahin, Karikó, and Türeci]{mrna-based_2014}
Ugur Sahin, Katalin Karikó, and Özlem Türeci.
\newblock {mRNA}-based therapeutics — developing a new class of drugs.
\newblock \emph{Nature Reviews Drug Discovery}, 13\penalty0 (10):\penalty0 759--780, 2014.
\newblock ISSN 1474-1784.
\newblock \doi{10.1038/nrd4278}.
\newblock URL \url{https://doi.org/10.1038/nrd4278}.

\bibitem[Schneider et~al.(2015)Schneider, Lowe, Sayle, and Landrum]{fingerprint}
Nadine Schneider, Daniel Lowe, Roger Sayle, and Gregory Landrum.
\newblock Development of a novel fingerprint for chemical reactions and its application to large-scale reaction classification and similarity.
\newblock \emph{Journal of chemical information and modeling}, 55, 02 2015.
\newblock \doi{10.1021/acs.jcim.5b00046}.

\bibitem[Segler et~al.(2017)Segler, Preu{\ss}, and Waller]{AlphaChem}
Marwin H.~S. Segler, Mike Preu{\ss}, and Mark~P. Waller.
\newblock Towards "alphachem": Chemical synthesis planning with tree search and deep neural network policies.
\newblock \emph{CoRR}, abs/1702.00020, 2017.
\newblock URL \url{http://arxiv.org/abs/1702.00020}.

\bibitem[Segler et~al.(2018)Segler, Preuss, and Waller]{segler_planning_2018}
Marwin H.~S. Segler, Mike Preuss, and Mark~P. Waller.
\newblock Planning chemical syntheses with deep neural networks and symbolic {AI}.
\newblock \emph{Nature}, 555\penalty0 (7698):\penalty0 604--610, 2018.
\newblock ISSN 1476-4687.
\newblock \doi{10.1038/nature25978}.
\newblock URL \url{https://doi.org/10.1038/nature25978}.

\bibitem[Shi et~al.(2021)Shi, Luo, Xu, and Tang]{Diffusion-1}
Chence Shi, Shitong Luo, Minkai Xu, and Jian Tang.
\newblock Learning gradient fields for molecular conformation generation, 2021.
\newblock URL \url{https://arxiv.org/abs/2105.03902}.

\bibitem[Silver et~al.(2016)Silver, Huang, Maddison, Guez, Sifre, van~den Driessche, Schrittwieser, Antonoglou, Panneershelvam, Lanctot, Dieleman, Grewe, Nham, Kalchbrenner, Sutskever, Lillicrap, Leach, Kavukcuoglu, Graepel, and Hassabis]{AlphaGo}
David Silver, Aja Huang, Chris~J. Maddison, Arthur Guez, Laurent Sifre, George van~den Driessche, Julian Schrittwieser, Ioannis Antonoglou, Veda Panneershelvam, Marc Lanctot, Sander Dieleman, Dominik Grewe, John Nham, Nal Kalchbrenner, Ilya Sutskever, Timothy Lillicrap, Madeleine Leach, Koray Kavukcuoglu, Thore Graepel, and Demis Hassabis.
\newblock Mastering the game of go with deep neural networks and tree search.
\newblock \emph{Nature}, 529\penalty0 (7587):\penalty0 484--489, 2016.
\newblock ISSN 1476-4687.
\newblock \doi{10.1038/nature16961}.
\newblock URL \url{https://doi.org/10.1038/nature16961}.

\bibitem[Silver et~al.(2017)Silver, Schrittwieser, Simonyan, Antonoglou, Huang, Guez, Hubert, Baker, Lai, Bolton, Chen, Lillicrap, Hui, Sifre, van~den Driessche, Graepel, and Hassabis]{AlphaZero}
David Silver, Julian Schrittwieser, Karen Simonyan, Ioannis Antonoglou, Aja Huang, Arthur Guez, Thomas Hubert, Lucas Baker, Matthew Lai, Adrian Bolton, Yutian Chen, Timothy Lillicrap, Fan Hui, Laurent Sifre, George van~den Driessche, Thore Graepel, and Demis Hassabis.
\newblock Mastering the game of go without human knowledge.
\newblock \emph{Nature}, 550\penalty0 (7676):\penalty0 354--359, 2017.
\newblock ISSN 1476-4687.
\newblock \doi{10.1038/nature24270}.
\newblock URL \url{https://doi.org/10.1038/nature24270}.

\bibitem[Simonovsky and Komodakis(2018)]{GraphVAE}
Martin Simonovsky and Nikos Komodakis.
\newblock Graphvae: Towards generation of small graphs using variational autoencoders, 2018.

\bibitem[Stern and {McNeil}(2007)]{stern_nanotechnology_safety_2007}
Stephan~T. Stern and Scott~E. {McNeil}.
\newblock Nanotechnology safety concerns revisited.
\newblock \emph{Toxicological Sciences}, 101\penalty0 (1):\penalty0 4--21, 2007.
\newblock ISSN 1096-6080.
\newblock \doi{10.1093/toxsci/kfm169}.
\newblock URL \url{https://doi.org/10.1093/toxsci/kfm169}.
\newblock \_eprint: https://academic.oup.com/toxsci/article-pdf/101/1/4/10976607/kfm169.pdf.

\bibitem[Sud et~al.(2007)Sud, Fahy, Cotter, Brown, Dennis, Glass, Merrill, Murphy, Raetz, Russell, and Subramaniam]{LipidMaps}
Manish Sud, Eoin Fahy, Deirdre Cotter, H.~Alex Brown, Edward~A. Dennis, Christopher~K. Glass, Alfred H.~Jr. Merrill, Robert~C. Murphy, Christian R.~H. Raetz, David~W. Russell, and Shankar Subramaniam.
\newblock L{M}{S}{D}: Lipid maps® structure database.
\newblock Nucleic Acids Research, 2007.
\newblock PMID: 17098933.

\bibitem[Sutton and Barto(2018)]{RL-textbook}
Richard~S Sutton and Andrew~G Barto.
\newblock \emph{Reinforcement learning: An introduction}.
\newblock MIT press, 2018.

\bibitem[Swanson et~al.(2024)Swanson, Liu, Catacutan, Arnold, Zou, and Stokes]{SyntheMol}
Kyle Swanson, Gary Liu, Denise~B. Catacutan, Autumn Arnold, James Zou, and Jonathan~M. Stokes.
\newblock Generative ai for designing and validating easily synthesizable and structurally novel antibiotics.
\newblock \emph{Nature Machine Intelligence}, 6:\penalty0 338--353, 2024.
\newblock \doi{10.1038/s42256-024-00809-7}.

\bibitem[Wan et~al.(2013)Wan, Zeiler, Zhang, Le~Cun, and Fergus]{dropout}
Li~Wan, Matthew Zeiler, Sixin Zhang, Yann Le~Cun, and Rob Fergus.
\newblock Regularization of neural networks using dropconnect.
\newblock In Sanjoy Dasgupta and David McAllester, editors, \emph{Proceedings of the 30th International Conference on Machine Learning}, volume~28 of \emph{Proceedings of Machine Learning Research}, pages 1058--1066, Atlanta, Georgia, USA, 17--19 Jun 2013. PMLR.
\newblock URL \url{https://proceedings.mlr.press/v28/wan13.html}.

\bibitem[Wei et~al.(2023)Wei, Sun, Cheng, Chatterjee, Traylor, Johnson, Coquelin, Wang, Torres, Lian, Wang, Xiao, Hodges, and Siegwart]{published-data-6}
Tuo Wei, Yehui Sun, Qiang Cheng, Sumanta Chatterjee, Zachary Traylor, Lindsay~T. Johnson, Melissa~L. Coquelin, Jialu Wang, Michael~J. Torres, Xizhen Lian, Xu~Wang, Yufen Xiao, Craig~A. Hodges, and Daniel~J. Siegwart.
\newblock Lung {SORT} {LNPs} enable precise homology-directed repair mediated {CRISPR}/cas genome correction in cystic fibrosis models.
\newblock \emph{Nature Communications}, 14\penalty0 (1):\penalty0 7322, 2023.
\newblock ISSN 2041-1723.
\newblock \doi{10.1038/s41467-023-42948-2}.
\newblock URL \url{https://doi.org/10.1038/s41467-023-42948-2}.

\bibitem[Wittrup et~al.(2015)Wittrup, Ai, Liu, Hamar, Trifonova, Charisse, Manoharan, Kirchhausen, and Lieberman]{ionizable-lipid-3}
Anders Wittrup, Angela Ai, Xing Liu, Peter Hamar, Radiana Trifonova, Klaus Charisse, Muthiah Manoharan, Tomas Kirchhausen, and Judy Lieberman.
\newblock Visualizing lipid-formulated {siRNA} release from endosomes and target gene knockdown.
\newblock \emph{Nature Biotechnology}, 33\penalty0 (8):\penalty0 870--876, 2015.
\newblock ISSN 1546-1696.
\newblock \doi{10.1038/nbt.3298}.
\newblock URL \url{https://doi.org/10.1038/nbt.3298}.

\bibitem[Xu et~al.(2021)Xu, Saltzman, and Piotrowski-Daspit]{ionizable-lipid-4}
Emily Xu, W.~Mark Saltzman, and Alexandra~S. Piotrowski-Daspit.
\newblock Escaping the endosome: assessing cellular trafficking mechanisms of non-viral vehicles.
\newblock \emph{Journal of Controlled Release}, 335:\penalty0 465--480, 2021.
\newblock ISSN 0168-3659.
\newblock \doi{https://doi.org/10.1016/j.jconrel.2021.05.038}.
\newblock URL \url{https://www.sciencedirect.com/science/article/pii/S0168365921002728}.

\bibitem[Xu et~al.(2024)Xu, Ma, Cui, Chen, Xu, Gong, Golubovic, Zhou, Wang, Varley, Lu, Wang, and Li]{AGILE}
Yue Xu, Shihao Ma, Haotian Cui, Jingan Chen, Shufen Xu, Fanglin Gong, Alex Golubovic, Muye Zhou, Kevin~Chang Wang, Andrew Varley, Rick Xing~Ze Lu, Bo~Wang, and Bowen Li.
\newblock {AGILE} platform: a deep learning powered approach to accelerate {LNP} development for {mRNA} delivery.
\newblock \emph{Nature Communications}, 15\penalty0 (1):\penalty0 6305, 2024.
\newblock ISSN 2041-1723.
\newblock \doi{10.1038/s41467-024-50619-z}.
\newblock URL \url{https://doi.org/10.1038/s41467-024-50619-z}.

\bibitem[Yang et~al.(2019)Yang, Swanson, Jin, Coley, Eiden, Gao, Guzman-Perez, Hopper, Kelley, Mathea, Palmer, Settels, Jaakkola, Jensen, and Barzilay]{Chemprop}
Kevin Yang, Kyle Swanson, Wengong Jin, Connor Coley, Philipp Eiden, Hua Gao, Angel Guzman-Perez, Timothy Hopper, Brian Kelley, Miriam Mathea, Andrew Palmer, Volker Settels, Tommi Jaakkola, Klavs Jensen, and Regina Barzilay.
\newblock Analyzing learned molecular representations for property prediction.
\newblock \emph{Journal of Chemical Information and Modeling}, 59\penalty0 (8):\penalty0 3370--3388, 2019.
\newblock \doi{10.1021/acs.jcim.9b00237}.
\newblock Correction published: J Chem Inf Model. 2019 Dec 23;59(12):5304-5305. doi: 10.1021/acs.jcim.9b01076.

\bibitem[Yu et~al.(2020)Yu, Liu, Cheng, Wei, Lee, Zhang, and Siegwart]{published-data-5}
Xueliang Yu, Shuai Liu, Qiang Cheng, Tuo Wei, Sang Lee, Di~Zhang, and Daniel~J. Siegwart.
\newblock Lipid-modified aminoglycosides for mrna delivery to the liver.
\newblock \emph{Advanced Healthcare Materials}, 9\penalty0 (7):\penalty0 1901487, 2020.
\newblock \doi{https://doi.org/10.1002/adhm.201901487}.
\newblock URL \url{https://onlinelibrary.wiley.com/doi/abs/10.1002/adhm.201901487}.

\end{thebibliography}

\appendix
\section{Lipid Building Block Dataset Construction and Selected Examples}
\label{sec:appendix-dataset-example}

We here discuss in detail how the lipid building block dataset construction process is performed and present selected examples of lipid building blocks.

For the ionizable lipid head dataset, we first filter ionizable molecules from the ZINC20 molecule dataset based on the following three criteria:
\begin{enumerate}
    \item Molecular weight < 500 g/mol.
    \item Log P < 0 where the log P is the logarithm (base 10) of the partition coefficient P, which is the ratio of the concentrations of a compound in a mixture of two immiscible phases: typically a hydrophobic solvent and water.
    \item Molecules with amine functional groups but not ammonium based molecules.
\end{enumerate}
Since the lipid head building blocks are expected to react with lipid tail building blocks to generate product lipids, we further filter reactive molecules from our ionizable molecule set. We identify reactive molecules by filtering whether or not the molecule contains certain functional groups that participate in common reactions. Specifically, we check whether our candidate molecule contains carboxyl group (i.e., -COOH), hydroxyl group (i.e., -OH), or amine group (i.e., -N or -NH or -NH2). If the candidate molecule contains any one or more of the target functional groups, we consider the molecule to be reactive. Our filtered reactive ionizable molecule set becomes our ionizable lipid head dataset.

In terms of lipid tail building block dataset, we again filter from the ZINC20 molecule dataset so that our lipid tail building blocks are purchasable. We identify a lipid tail building block if the molecule is similar to an actual lipid tail (i.e., the tail component of an existing lipid). We use a lipid dataset sourced from the LipidMaps database and the SwissLipid database \citep{LipidMaps, SwissLipids}. Leveraging the Lipid Analyzer, an implemented toolkit for lipid tail extraction, we extract lipid tails from this lipid dataset. The Lipid Analyzer finds the lipid head for a given lipid, we then extract lipid tails by removing the head substructure. For a given molecular structure, the algorithm first recognizes ring structures. For all possible arrangements of the rings, the algorithm then removes carbon atoms that are not near any hydrophilic atom and not forming any ring. If only one substructure remains after the operation, and if the log P value of this substructure is low enough, this substructure will be identified as the lipid head. It's then easy to extract tail substructures by removing the head substructure.

We then conduct a similarity comparison to filter those molecules from the ZINC20 dataset that are similar to a real lipid tail. The similarity is measured by the Graph Edit Distance (GED) between two molecular structures, quantifying the minimum number of operations required to transform one graph into another. We only select molecules that has GED value smaller or equal to 1 with a real lipid tail. Meanwhile, we also consider the similarity between molecular fingerprint representations, Daylight fingerprint and Extended Connectivity Fingerprint 4 (ECFP4) are considered. The resulting molecules make up our lipid tail building block dataset.

\begin{figure}
    \centering
    \includegraphics[width=0.8\linewidth]{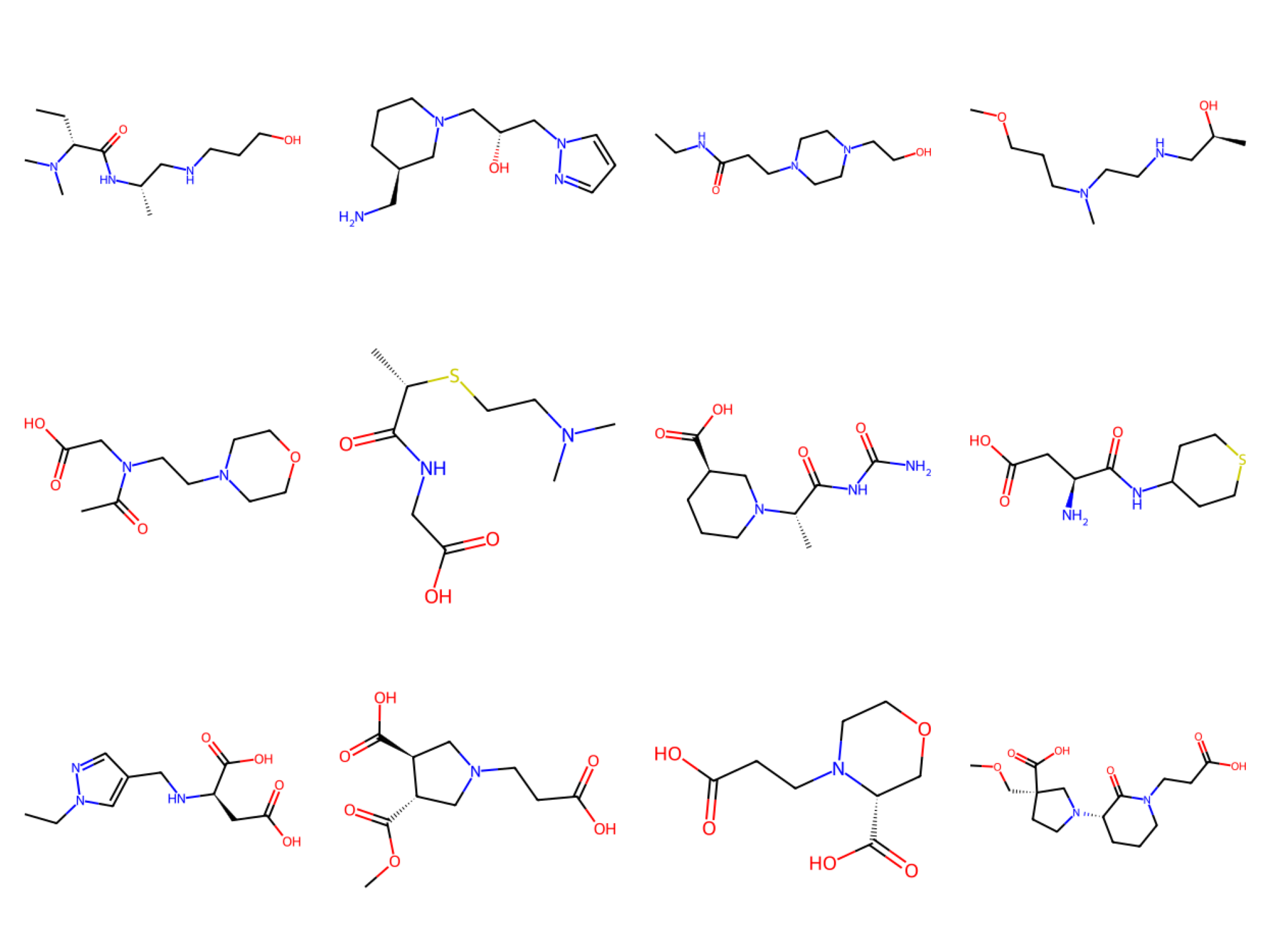}
    \caption{Selected examples of lipid head building blocks.}
    \label{fig:example-lipid-head}
\end{figure}

Selected examples of our lipid head building blocks with different functional groups are shown in Figure \ref{fig:example-lipid-head}. The first row presents examples with no carboxyl group, one hydroxyl group, and two amine groups; the second row presents examples with one carboxyl group, no hydroxyl group, and one amine group; the third row presents examples with two carboxyl groups, no hydroxyl group, and one amine group. Note that we only consider independent amine groups that are linked to only simple carbon atoms. For example, in the first example of the first row, we find a secondary amine group (i.e., -NH) linked to a carbonyl group (i.e., C=O, a carbon atom double-bonded to an oxygen atom). The linkage with the carbonyl group forms a more complicated substructure, which influences the reactive performance of the amine group. We therefore exclude this secondary amine group when counting our target functional groups. Similarly, the nitrogen-nitrogen bond (i.e., N-N) which appears in a five-membered nitrogen-containing ring influences the reactive property of the nitrogen, and we also exclude these nitrogen atoms when counting the occurrence of amine functional groups.

\begin{figure}
    \centering
    \includegraphics[width=0.8\linewidth]{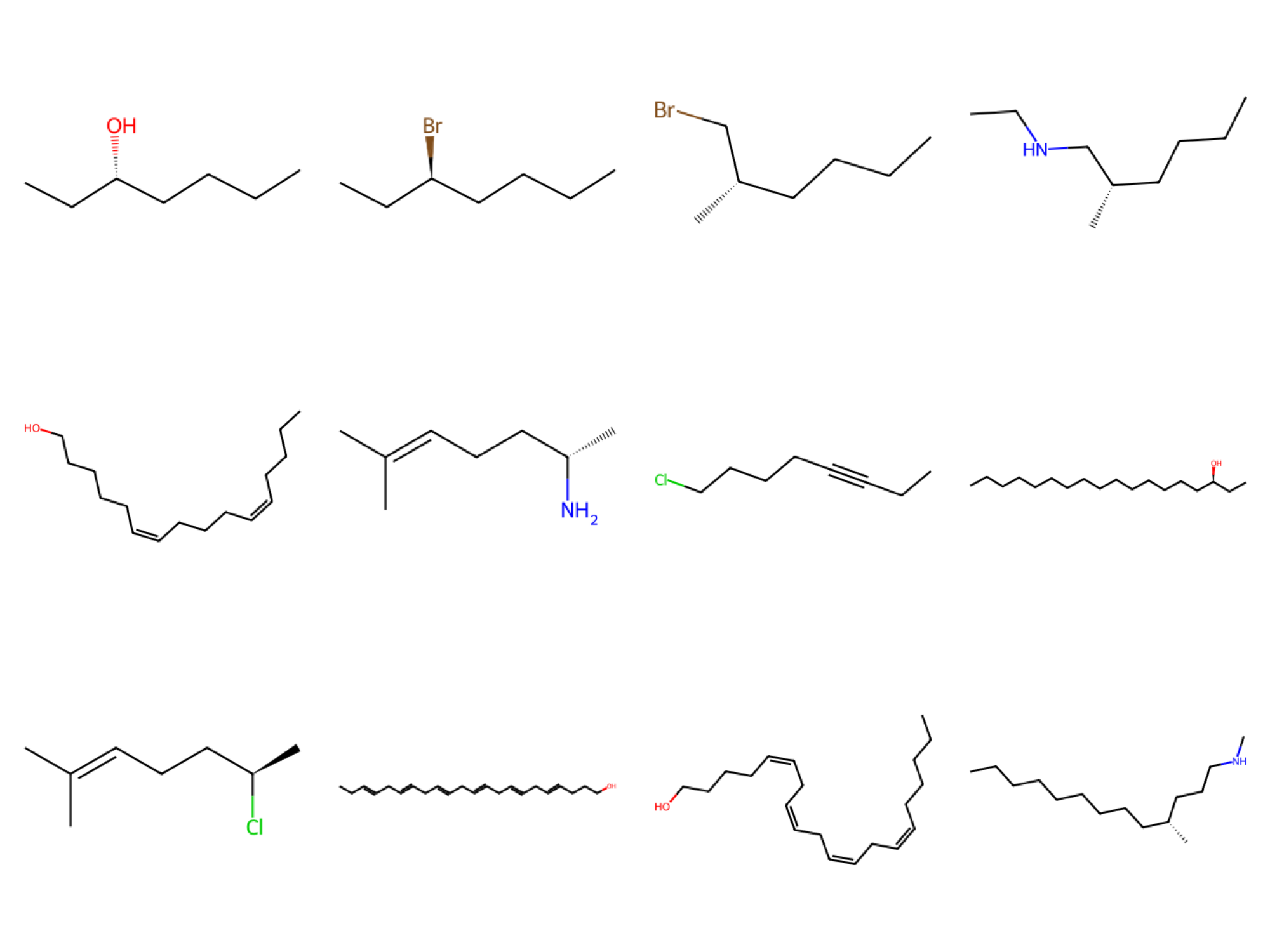}
    \caption{Selected examples of lipid tail building blocks.}
    \label{fig:example-lipid-tail}
\end{figure}

Figure \ref{fig:example-lipid-tail} shows selected examples of lipid tail building blocks. As we can see, a lipid tail usually consists of a carbon chain and a functional group. The functional group may be hydroxyl group, amine group, or a halogen atom.

\section{SyntheMol for Lipid Generation}
\label{sec:appendix-synthemol}

\begin{figure}
    \centering
    \includegraphics[width=\linewidth]{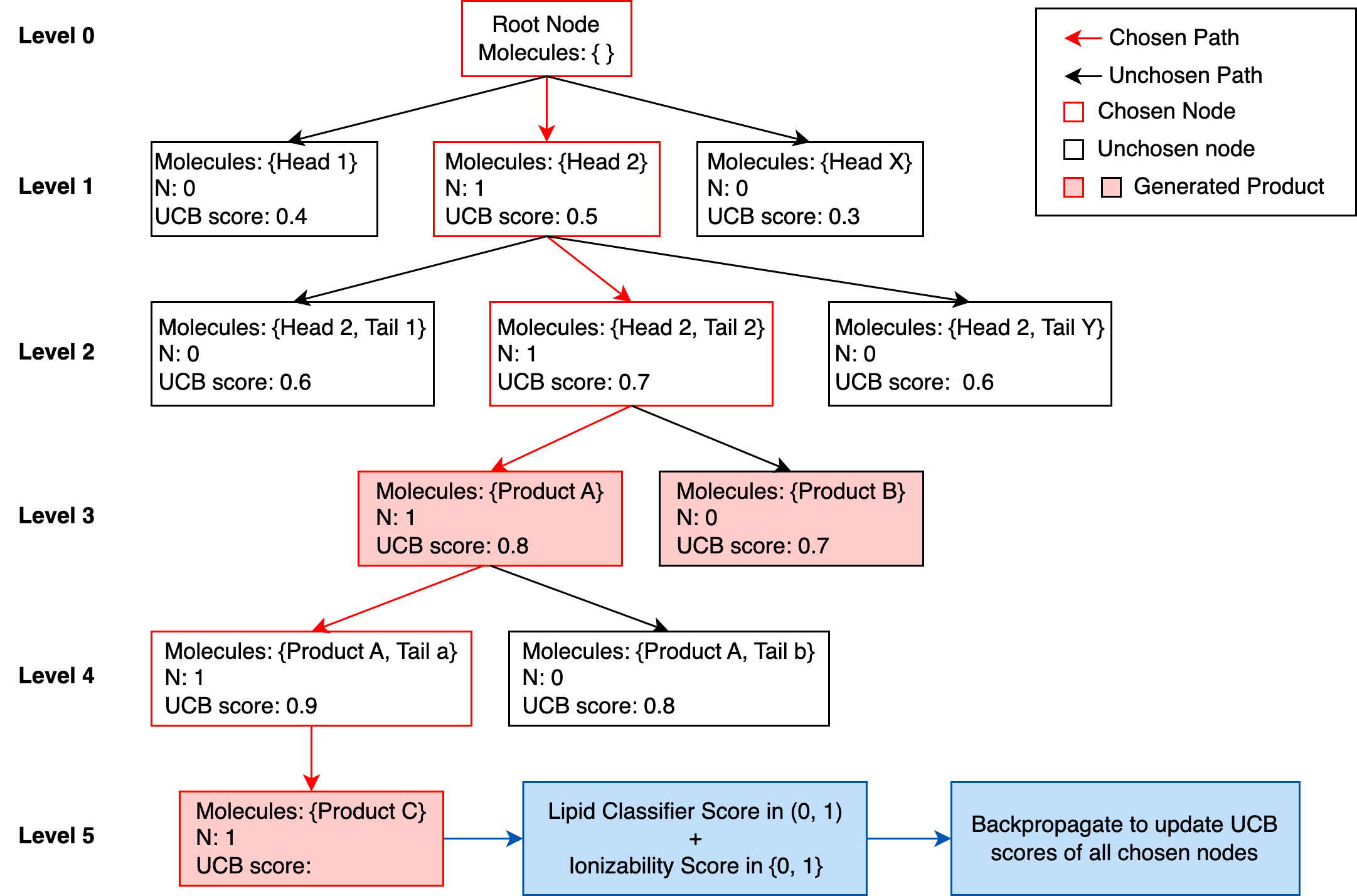}
    \caption{SyntheMol for lipid generation.}
    \label{fig:NaiveMCTS}
\end{figure}

Figure \ref{fig:NaiveMCTS} illustrates a typical simulation of the MCTS process in our lipid generation. In the search tree, each node stores crucial statistics: the molecules represented by the node, the node's visit count $N$, and the Upper Confidence Bound (UCB) score. The UCB score guides the node selection process using the UCB action selection criterion \citep{RL-textbook, UCT}. The UCB score combines an action value $Q$, which represents the average of the total action values $W$ (i.e., the sum of the values of the final products passing through the node), with an upper confidence term $U$. This term is defined as $U(node) \propto \frac{P(node)}{1 + N(node)}$, where $P(node)$ indicates the property scores assigned to the node. The complete formula for the UCB score is as follows:
\begin{align}
UCB\_score(node) &= Q(node) + U(node) \\
&= \frac{W(node)}{N(node)} + c \cdot P(node) \cdot \frac{\sqrt{1 + n}}{1 + N(node)}
\label{eq:naive-mcts-ucb}
\end{align}
where $n$ is the total visit count of all nodes at the same level as the target node and $c$ is a constant exploration parameter controlling the level of exploration. Note that the $P$ value is the property score calculated by our property predictors. When a node represents multiple molecules, the $P$ value for that node is computed as the average of the property scores for each constituent molecule.

It is important to note that the assignment of property scores to nodes at non-terminal levels—representing either building block molecules or intermediates—may initially seem illogical, as direct evaluation of these entities' properties does not typically yield meaningful insights. However, this approach does not compromise the long-term efficacy of the algorithm. Over time, as more product molecules are synthesized and action values refined, the utility of early-stage evaluations is validated. SyntheMol's findings corroborate this approach, demonstrating that despite low scores of individual building blocks, the synthesized molecules often exhibit significantly higher scores, which are effectively identified by MCTS, highlighting its capability to uncover promising molecules overlooked by simpler scoring methods \citep{SyntheMol}.

The MCTS algorithm commences at a root node (level 0), an empty initial node, which is then expanded with child nodes representing available lipid head building blocks defined within our chemical space (level 1). Each child node receives an UCB score, with the node exhibiting the highest UCB score selected for further expansion. This selected node is expanded to include a lipid tail building block, generating second-level nodes where each combination represents potential reactants (level 2).

Subsequently, the node with the highest UCB score from level 2 is selected and expanded. This stage differs from prior expansions in that it now entails actual chemical reactions between the two building blocks within the node. The resulting third-level nodes embody all conceivable products predicted by forward reaction mechanisms (level 3). This iterative expansion continues until either a valid final product (i.e., a two-tail lipid in our case) is synthesized, or further reactions become untenable with the existing building blocks in our dataset.

Upon terminating the simulation with a valid final product, the synthesized molecule is assessed using our property predictors. The resultant property value is then backpropagated to update the UCB scores along the simulated pathway. For analysis purpose, we meticulously document all products, including intermediates, generated during each simulation.

\section{Algorithm of Guided Monte Carlo Tree Search for Lipid Generation}
\label{sec:appendix-algo-guided-mcts}

We define a \texttt{Node} class to represent the nodes appeared in the tree search. The \texttt{Node} class has the following attributes:
\begin{itemize}
    \item \texttt{state}: Stores the SMILES representation of the molecule represented by this node.
    \item \texttt{N}: A visit count of the node, initialized to zero.
    \item \texttt{P}: The prior probability assigned based on predictions from a policy network.
    \item \texttt{W}: The cumulative value sum, representing the total assessed value of this node's state.
    \item \texttt{children}: A dictionary to hold child nodes, with the keys of the dictionary to be actions (i.e., the next building blcoks to select) and the values to be corresponding child nodes.
\end{itemize}

\begin{algorithm}
\caption{Policy Network Guided MCTS}
\begin{algorithmic}[1]
\Require Product(): reaction predictor
\Require PropertyScore(): property score predictor
\Require SearchProbability(): search probability calculator

\State $f_\theta \gets \text{randomly initialized neural network with parameter } \theta$
\State $\lambda \gets \text{regularization constant}$ 
\State $\alpha \gets \text{learning rate}$ 
\State $c \gets \text{exploration weight}$ 
\State $D = \emptyset$
\Statex
\For{\textbf{each iteration}}
    \For{\textbf{each play}}
        \State $S_0 \gets \text{empty root state}$
        \State $D_{play} \gets \Call{MCTS}{S_0}$
        \State $ D = D \cup D_{play}$
    \EndFor
    \For{\textbf{each epoch}}
    \State Train $f_\theta$ using $D$
    \EndFor
    \State Reset $D = \emptyset$
\EndFor
\Statex

\Function{MCTS}{$root\_state$}
    \State Initialize $root\_node$ with $root\_state$
    \State \Call{Expand}{$root\_node$}
    \State $generation = \emptyset$
    \For{\textbf{each simulation}}
        \State $leaf\_node, search\_path \gets \Call{Select}{root\_node}$
        \If{$leaf\_node$ is two-tail lipid}
            \State $v$ = PropertyScore($leaf\_node$)
            \State Add $search\_path$ to $generation$
        \Else
            \State \Call{Expand}{$leaf\_node$}
            \State $v \gets \Call{Rollout}{leaf\_node}$
        \EndIf
        \State \Call{Backpropagate}{$v$, $search\_path$}
    \EndFor
    \State Write $generation$ to log file
    \State \Return visit counts of all state-action pairs
\EndFunction
\Statex
\end{algorithmic}
\end{algorithm}

\begin{algorithm}
\caption{Select Function in MCTS}
\begin{algorithmic}[1] 
\Function{Select}{$root\_node$}
    \State $search\_path = [\ ]$
    \State $node \gets root\_node$
    \While{$node.children$ is not empty}
        \State $selected\_action = \arg\max_a \Call{UCB\_Score}{node, node.children[a]}$ 
        \State $node \gets node.children[selected\_action]$ 
        \State Add $node$ to $search\_path$
    \EndWhile
    \State Update $node.state$ with Product($node.state, selected\_action$)
    \State \Return $node$, $search\_path$
\EndFunction
\end{algorithmic}
\end{algorithm}

\begin{algorithm}
\caption{Expand Function in MCTS}
\begin{algorithmic}[1] 
\Function{Expand}{$node$}
    \State $A(node) \gets \Call{next\_building\_blocks}{node}$
    \State $P(node, \cdot) = f_\theta (node.state, A(node))$
    \For{$a$ in $A(node)$}
        \State Initialize $child\_node$ with $P(node, a)$
        \State $node.children[a] = child\_node$
    \EndFor
\EndFunction
\end{algorithmic}
\end{algorithm}

\begin{algorithm}
\caption{Backpropagate Function in MCTS}
\begin{algorithmic}[1] 
\Function{Backpropagate}{$v$, $search\_path$}
    \For{$node$ in $search\_path$}
        \State $node.N \gets node.N + 1$
        \State $node.W \gets node.W + v$
        \State $root\_node.N \gets root\_node.N + 1$
    \EndFor
\EndFunction
\end{algorithmic}
\end{algorithm}

\begin{algorithm}
\caption{Rollout Function in MCTS}
\begin{algorithmic}[1] 
\Function{Rollout}{$node$}
    \While{$node.state$ is not two-tail lipid}
        \State $A(node) \gets \Call{next\_building\_blocks}{node}$
        \State \Return 0 if $A(node)$ is empty
        \State $a \gets$ a random choice from $A(node)$
        \State $node \gets$ new node with Product($node.state$, $a$)
    \EndWhile
    \State \Return PropertyScore($node$)
\EndFunction
\end{algorithmic}
\end{algorithm}

\begin{algorithm}
\caption{UCB Score Calculation Function in MCTS}
\begin{algorithmic}[1] 
\Function{UCB\_Score}{$parent\_node$, $child\_node$}
    \If{$child\_node.N = 0$}
        \State $Q = 0$
    \Else
        \State $Q = \frac{child\_node.W}{child\_node.N}$
    \EndIf
    \State $U = c \cdot child\_node.P \cdot \frac{\sqrt{parent\_node.N}}{child\_node.N + 1}$
    \State \Return $Q + U$
\EndFunction
\end{algorithmic}
\end{algorithm}

\begin{algorithm}
\caption{Get Action Space Function in MCTS}
\begin{algorithmic}[1] 
\Require $max\_expand\_num$ a pre-determined number
\Function{next\_building\_blocks}{$node$}
    \If{$node$ \text{is empty node}}
        \State $A(node) \gets \text{$max\_expand\_num$ random samples from head search space}$
    \Else
        \State $reactive\_tail\_set \gets \text{tails which can react with } node.state$
        \State $A(node) \gets \text{$max\_expand\_num$ random samples from $reactive\_tail\_set$}$
    \EndIf
    \State \Return $A(node)$
\EndFunction

\end{algorithmic}
\end{algorithm}

\newpage
\section{Policy Network Training}
\label{chapter:appendix-policy-network-training}

As is mentioned before, the visit counts of all state-action pairs that appear in the tree search are recorded to be the training data of the policy network. However, the search data we collect are highly imbalanced in the sense that there will be a lot more state-action pairs in later levels. The number of state-action pairs from the first level (i.e., when state is the empty state and actions are the chosen head building blocks) is very limited, being the number of different head building blocks that are expanded by the root node. Meanwhile, search data from later levels will be very sparse in the sense that the majority of the state-action pairs will have zero visit count. In order to better utilize our limited data, we propose a customized method for training the policy network.

Let $f(s, a)$ denote the naive policy network output (before softmax) for the state-action pair $(s, a)$, $p(s, a)$ denote the corresponding predicted prior (which will be used in the UCB score calculation), and $\pi (s, a)$ be the search probability. We here adopt the search probability definition as proposed in the AlphaZero algorithm \citep{AlphaZero}.
\begin{equation}
    p(s, a_1) = \frac{e^{f(s, a_1)}}{\sum_a e^{f(s, a)}}
\end{equation}
\begin{equation}
    \pi(s, a_1) = \frac{N(s, a_1)^{\frac{1}{\tau}}}{\sum_a N(s, a)^{\frac{1}{\tau}}}
\end{equation}
where $\tau$ is a temperature parameter controlling the level of exploration. When $\tau$ tends to infinity, the search probability is the same as random selections. When $\tau$ is small, say, when $\tau = 1$, the search probability strictly follows the actual visit count distribution. The objective of the policy network training is to maximize the similarity of $p(s, a)$ and $\pi (s, a)$.

Now, suppose we take two state-action pairs, $(s, a_1)$ and $(s, a_2)$, at a time. We apply the log-ratio and get:
\begin{equation}
    \log \frac{p(s, a_1)}{p(s, a_2)} = f(s, a_1) - f(s, a_2)
\end{equation}
\begin{equation}
    \log \frac{\pi(s, a_1)}{\pi(s, a_2)} = \frac{1}{\tau} \log N(s, a_1) - \frac{1}{\tau} \log N(s, a_2)
\end{equation}
The previous objective is equivalent to maximize the similarity of $f(s, a_1) - f(s, a_2)$ and $\frac{1}{\tau} \log N(s, a_1) - \frac{1}{\tau} \log N(s, a_2)$.

We therefore customize the loss function to be the error between $f(s, a_1) - f(s, a_2)$ and their corresponding $\frac{1}{\tau} \log N(s, a_1) - \frac{1}{\tau} \log N(s, a_2)$, for any two state-action pairs, $(s, a_1)$ and $(s, a_2)$. This can be considered as a regression problem, MAE or MSE loss may be applied.

\section{Selected Examples of Synthesis Planning}
\label{chapter:appendix-synthesis-example}

\begin{figure}
    \centering
    \includegraphics[width=0.7\linewidth]{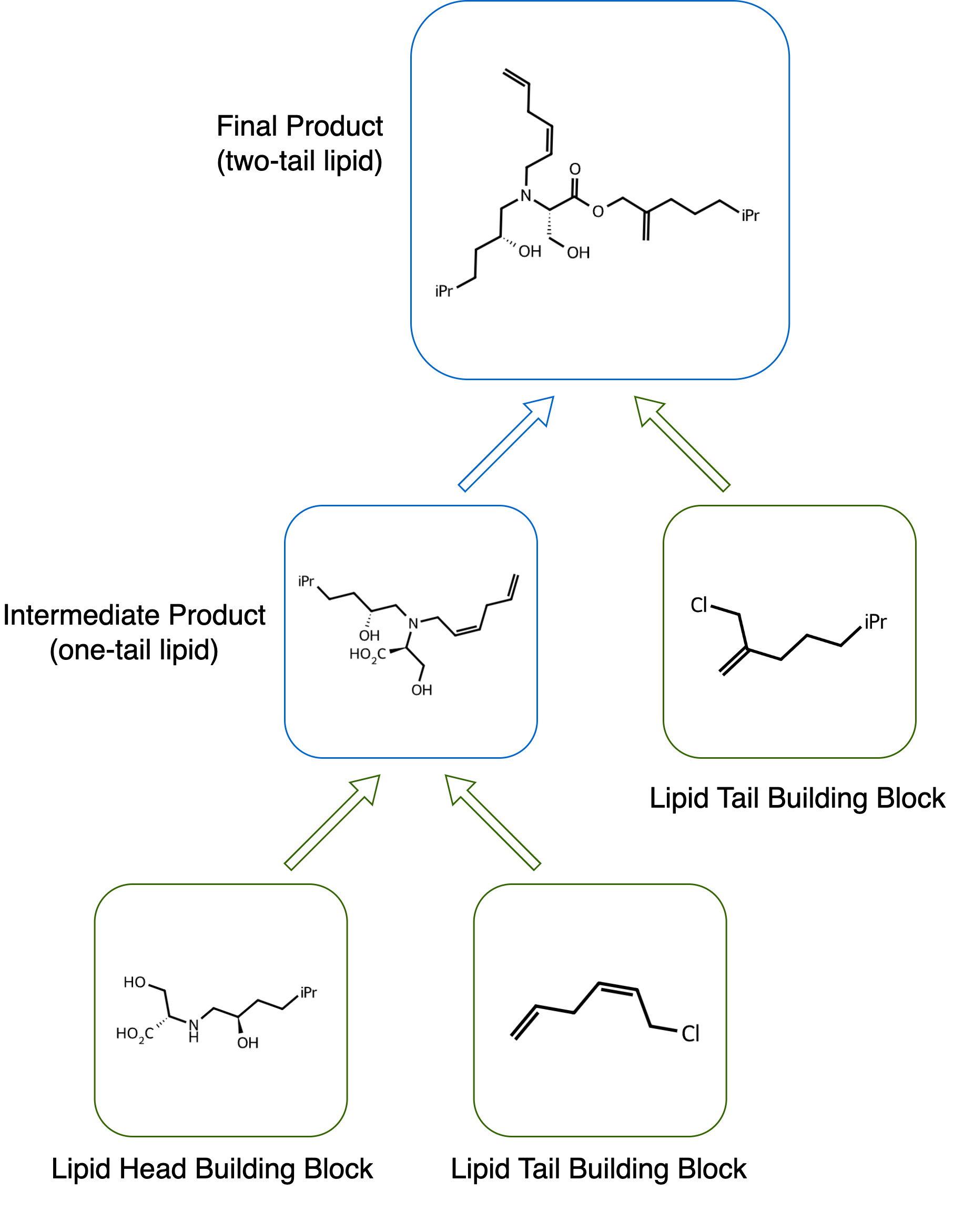}
    \caption{Example 1 of generated ionizable lipid with synthesis path.}
    \label{fig:example-1}
\end{figure}

\begin{figure}
    \centering
    \includegraphics[width=0.7\linewidth]{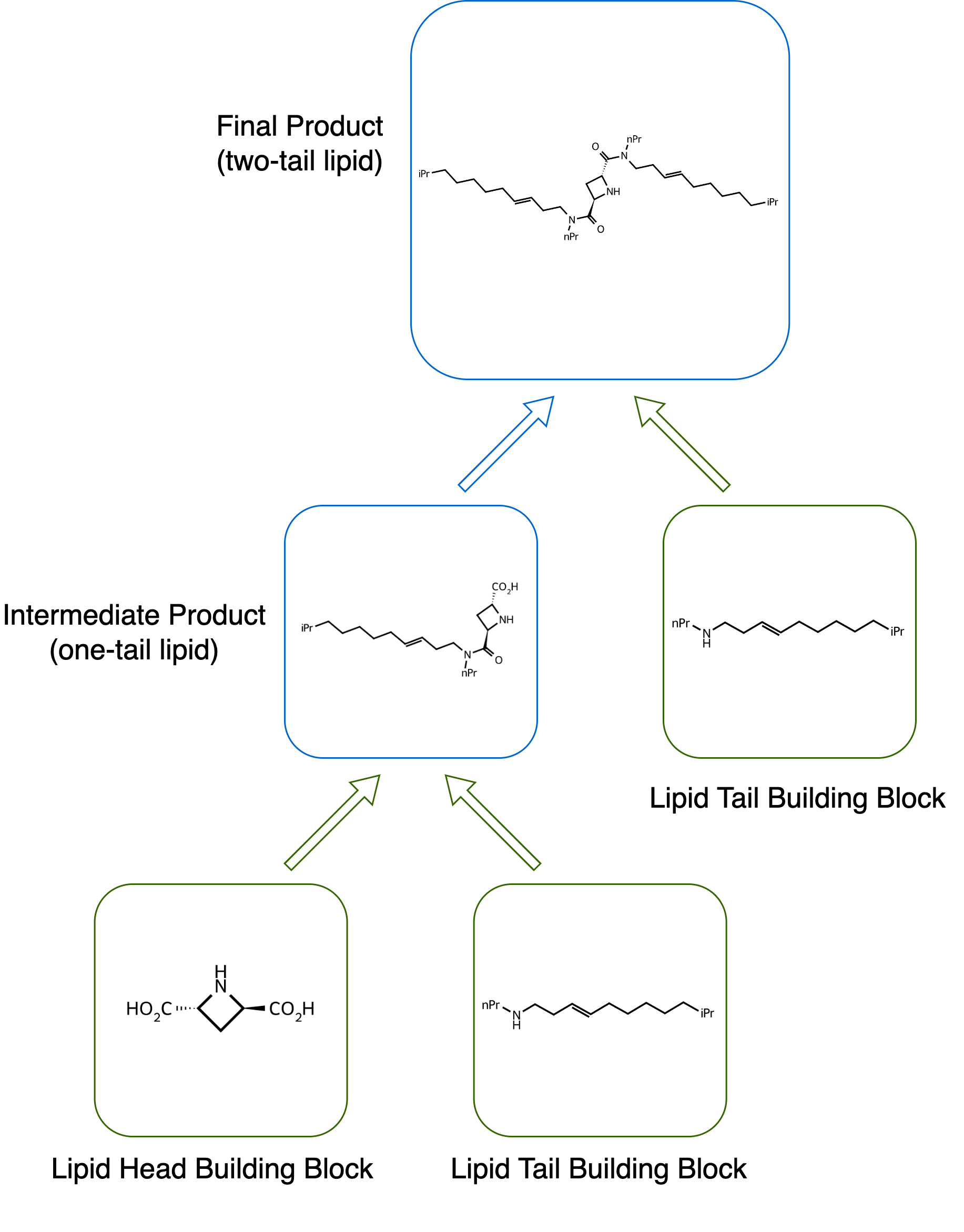}
    \caption{Example 2 of generated ionizable lipid with synthesis path.}
    \label{fig:example-2}
\end{figure}

Figures \ref{fig:example-1} and \ref{fig:example-2} give two examples of our generated ionizable lipids with their corresponding synthesis paths. Note that these synthesis paths have been validated by the retrosynthesis tool Syntheseus \citep{syntheseus}. In the figures below, the blue nodes represent product molecules (i.e., intermediate product or final product) and the green nodes represent building block molecules (i.e., lipid head building block or lipid tail building block).

\end{document}